\def\paperID{0000}
\def\confName{CVPR}
\def\confYear{2025}

\def\paperTitle{Cross-Self KV Cache Pruning for Efficient Vision-Language Inference}

\def\authorBlock{
    Xiaohuan Pei  \qquad
    Tao Huang  \qquad
    Chang Xu  \\
    School of Computer Science, Faculty of Engineering, The University of Sydney \\
    {\tt\small \{xiaohuan.pei, t.huang, c.xu\}@sydney.edu.au}
}

\newif\ifreview 
\newif\ifarxiv \newcommand{\arxiv}{\arxivtrue}
\newif\ifcamera 
\newif\ifrebuttal 

\arxiv

\pdfoutput=1
\documentclass[10pt,twocolumn,letterpaper]{article}
\ifreview \usepackage[review]{cvpr} \fi
\ifarxiv \usepackage[pagenumbers]{cvpr} \fi
\ifrebuttal \usepackage[rebuttal]{cvpr} \fi
\ifcamera \usepackage{cvpr} \fi

\usepackage{graphicx}	
\usepackage{amsmath}	
\usepackage{amssymb}	
\usepackage{booktabs}
\usepackage{times}
\usepackage{microtype}
\usepackage{epsfig}
\usepackage{caption}
\usepackage{float}
\usepackage{placeins}
\usepackage{color, colortbl}
\usepackage{stfloats}
\usepackage{enumitem}
\usepackage{tabularx}
\usepackage{xstring}
\usepackage{multirow}
\usepackage{xspace}
\usepackage{url}
\usepackage{subcaption}
\usepackage{xcolor}
\usepackage[hang,flushmargin]{footmisc}

\ifcamera \usepackage[accsupp]{axessibility} \fi

% ...

\ifarxiv  \fi

\newcommand{\R}[1]{{%
    \textbf{%
        \ifstrequal{#1}{1}{\textcolor{red}{R#1}}{%
        \ifstrequal{#1}{2}{\textcolor{blue}{R#1}}{%
        \ifstrequal{#1}{3}{\textcolor{magenta}{R#1}}{%
        \ifstrequal{#1}{4}{\textcolor{teal}{R#1}}{%
                           \textcolor{cyan}{R#1}%
        }}}}%
    }%
}}

\usepackage{xr-hyper}

\makeatletter
\newcommand*{\addFileDependency}[1]{
  \typeout{(#1)}
  \@addtofilelist{#1}
  \IfFileExists{#1}{}{\typeout{No file #1.}}
}

\makeatother

\definecolor{cvprblue}{rgb}{0.21,0.49,0.74}
\usepackage[pagebackref,breaklinks,colorlinks,allcolors=cvprblue]{hyperref}
\usepackage[capitalize]{cleveref}
\crefname{section}{Sec.}{Secs.}
\crefname{table}{Table}{Tables}
\crefname{figure}{Fig.}{Figs.}

\ifarxiv \crefname{appendix}{App.}{Apps.}
\else \crefname{appendix}{Suppl.}{Suppls.} \fi

\usepackage{multirow}
\usepackage{arydshln} 
\usepackage{xcolor}
\usepackage{algorithmicx}
\usepackage{algpseudocode}
\usepackage{algorithm}
\definecolor{mygray}{gray}{0.9} 
\usepackage{graphicx}
\usepackage{float}

\begin{document}
\title{\paperTitle}
\author{\authorBlock}
\maketitle

\begin{abstract}
KV cache pruning has emerged as a promising technique for reducing memory and computation costs in long-context auto-regressive generation. Existing methods for vision-language models (VLMs) typically rely on self-attention scores from large language models (LLMs) to identify and prune irrelevant tokens. However, these approaches overlook the inherent distributional discrepancies between modalities, often leading to inaccurate token importance estimation and the over-pruning of critical visual tokens. To address this, we propose decomposing attention scores into intra-modality attention (within the same modality) and inter-modality attention (across modalities), enabling more precise KV cache pruning by independently managing these distinct attention types. Additionally, we introduce an n-softmax function to counteract distribution shifts caused by pruning, preserving the original smoothness of attention scores and ensuring stable performance. Our final training-free method, \textbf{C}ross-\textbf{S}elf \textbf{P}runing (CSP), achieves competitive performance compared to models with full KV caches while significantly outperforming previous pruning methods. Extensive evaluations on MileBench, a benchmark encompassing 29 multimodal datasets, demonstrate CSP's effectiveness, achieving up to a 41\% performance improvement on challenging tasks like conversational embodied dialogue while reducing the KV cache budget by 13.6\%. 
The code is available at \href{https://github.com/TerryPei/CSP}{TerryPei/CSP}.
\end{abstract}
\section{Introduction}
\label{sec:intro}

The success of large language models (LLMs) \cite{achiam2023gpt, yang2023baichuan, touvron2023llama, bai2023qwen, bi2024deepseek, young2024yi} has propelled the advancement of large vision-language models (VLMs)~\cite{liu2024visual, liu2023improvedllava, team2023gemini, Qwen-VL, chen2023internvl, li2024mini, ye2023mplug}, enabling powerful integration and reasoning over multimodal inputs that combine both text and visual tokens. Unlike single-modal contexts, multimodal samples often comprise numerous images alongside text instructions, creating extended context lengths that challenge efficient inference. 


\begin{figure}[t]
    \centering
    \begin{subfigure}[b]{0.49\linewidth}
        \centering
        \includegraphics[height=3.5cm]{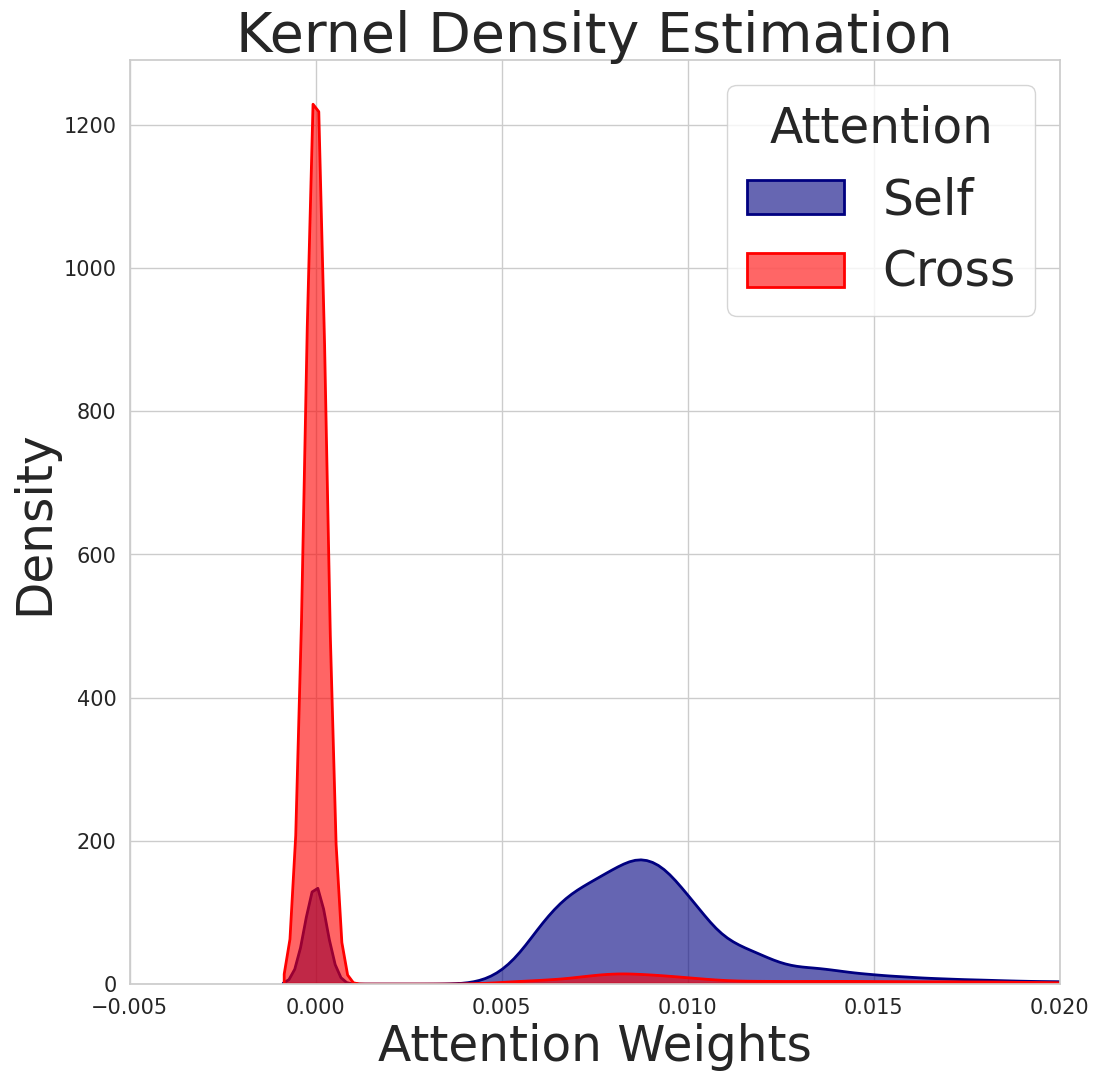} 
        \caption{KDE for Attention Map}
        \label{fig:kde}
    \end{subfigure}
    \hfill
    \begin{subfigure}[b]{0.49\linewidth}
        \centering
    \includegraphics[height=3.5cm]{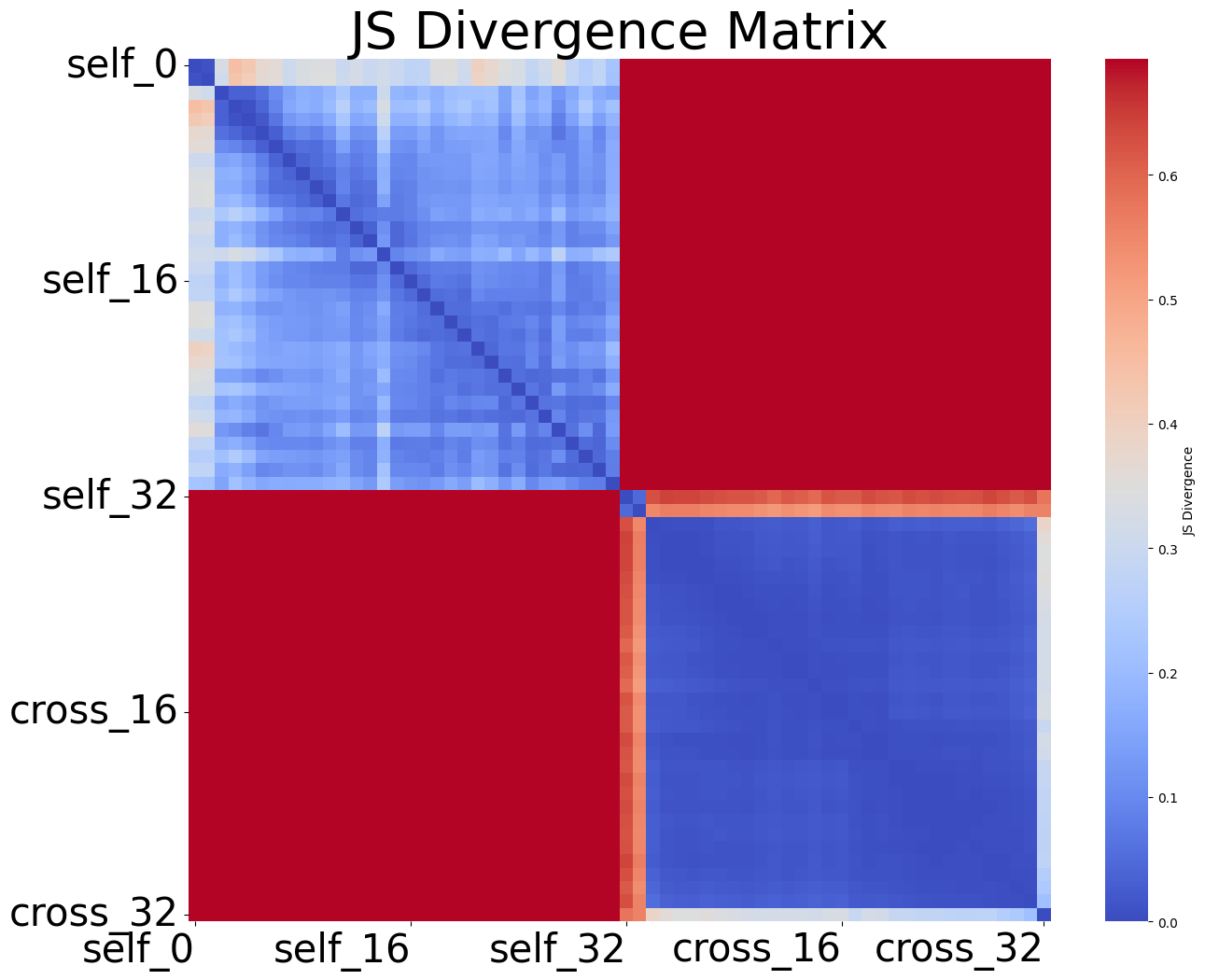} 
        \caption{Layer-Wise JS Estimations}
        \label{fig:js}
    \end{subfigure}
    \caption{Distribution gap between self-attention and cross-attention during the decoding process in VLM tasks: (a) Kernel Density Estimation (KDE) of the attention weight distributions, and (b) Jensen-Shannon (JS) divergence scores between cross-attention and self-attention across all layers.
    }
    \label{fig:1}
    \vspace{-5mm}
\end{figure}

To address the challenges of long-context generation, KV caching has become a standard technique, where previously computed keys and values in the attention layers are stored in memory for reuse during subsequent generation steps. However, this approach still faces significant memory limitations, particularly for GPU and machine memory constraints. Recent works \cite{li2024snapkv, zhang2024h2o, ren2024efficacy, liu2024minicache, cai2024pyramidkv} have explored pruning unimportant tokens within the KV cache to alleviate memory demands, primarily by leveraging attention scores. Methods such as SnapKV \cite{li2024snapkv} and H2O \cite{zhang2024h2o} apply this strategy to vision-language modeling (VLM) tasks by treating visual and text tokens uniformly across long sequences during pruning. Unfortunately, these methods rely on original attention scores that mix different modalities, potentially leading to suboptimal pruning outcomes.



In this paper, we identify a critical limitation in previous KV cache pruning methods: the distributional discrepancy between visual and textual modalities leads to inaccurate token importance estimation. Specifically, as illustrated in Figure \ref{fig:kde}, self-attention scores (within a single modality) and cross-attention scores (across modalities) exhibit distinct and non-overlapping distributions. This divergence highlights that each attention type captures unique aspects of the input space, reflecting modality-specific priorities during decoding. Furthermore, as shown in Figure \ref{fig:js}, the Jensen-Shannon (JS) divergence between cross-attention and self-attention distributions reveals substantial variation across layers in LLaVA-7b. Relying solely on these mixed distributions for pruning introduces a selection bias: the pruned tokens tend to cluster within a single region or modality. This imbalance disrupts cross-modal interactions, ultimately degrading the model's performance.

Inspired by these observations, we hypothesize that independently selecting important tokens from distinct distributions can provide a more balanced and effective pruning strategy. To achieve this, we propose decomposing the attention matrix into intra-modality attention (within the same modality) and inter-modality attention (across different modalities), and performing token selection independently within each category. This approach ensures accurate preservation of critical tokens from both modalities while maintaining the efficiency and simplicity of prior methods. Additionally, we observe that pruning inherently shifts the distribution of attention scores, leading to degraded performance. To mitigate this, we introduce an n-softmax function that smooths the post-pruning distribution, effectively restoring the original smoothness of the attention scores and improving overall performance.

Our final method, \textbf{C}ross-\textbf{S}elf \textbf{P}runing (CSP), effectively addresses the challenges of KV cache pruning by leveraging independent intra-modality and inter-modality token selection along with n-softmax smoothing. CSP achieves a superior balance between performance and memory efficiency, significantly reducing the KV cache size while preserving or even enhancing model accuracy. We conduct extensive experiments on various VLMs, including LLaVA-v1.5-7b \cite{liu2023improvedllava}, InternVL \cite{chen2023internvl}, and MobileVLM  \cite{chu2024mobilevlm}. The results demonstrate that CSP consistently outperforms existing methods like SnapKV \cite{li2024snapkv} and H2O \cite{zhang2024h2o}, achieving up to a 41\% improvement in performance on complex tasks such as conversational embodied dialogue 
\cite{shridhar2020alfred} , while reducing the KV cache budget by up to 13.6\%.




\section{Related Work}
\label{sec:related}

\subsection{Vision-language models}
Following the remarkable success of large language models (LLMs) \cite{achiam2023gpt, yang2023baichuan, touvron2023llama, bai2023qwen, bi2024deepseek, young2024yi}, recent research has focused on generative large vision-language models (VLMs) \cite{liu2024visual, liu2023improvedllava, team2023gemini, Qwen-VL, chen2023internvl, li2024mini, ye2023mplug} to enhance multimodal comprehension and generation by leveraging the generalization capabilities of LLMs. Using the multi-modal pre-trained visual foundation models such as CLIP~\cite{radford2021learning} as the visual encoder, existing methods commonly utilize extensive image-text data to align the visual encoder with LLMs, enabling the LLM to process and interpret visual inputs. For example, Flamingo~\cite{alayrac2022flamingo} incorporates visual features into the LLM through gated attention, while LLaVA~\cite{liu2024visual} connects the vision encoder and LLM via MLPs, demonstrating strong performance in multimodal dialogues.

\subsection{KV cache optimization}
Recent advancements in large language models (LLMs) have achieved notable success in optimizing KV cache for efficient long-context processing. 
Existing work on KV cache optimization \cite{li2024snapkv, zhang2024h2o, ren2024efficacy, adnan2024keyformer, liu2024minicache, cai2024pyramidkv, ge2023model} primarily utilizes attention scores to retain important tokens and improve memory efficiency in long-context processing. For example, SnapKV \cite{li2024snapkv} introduces a technique for identifying attention allocation patterns and compressing the KV cache by pooling key tokens. H2O \cite{zhang2024h2o} presents a dynamic eviction policy that balances recent and frequently accessed tokens, identifying heavy hitters based on attention scores alone. ReCo \cite{ren2024efficacy} improves existing eviction strategies through refined importance scoring and structured eviction scopes. Keyformer \cite{adnan2024keyformer} introduces gumbel softmax to relieve the impact of pruning tokens. More recent works \cite{wan2024look, zhang2024sparsevlm} have shifted towards adapting KV cache inference to multimodal contexts, where modality-specific properties are considered. For instance,  LOOK-M \cite{wan2024look} applies modality awareness to selectively evict image tokens and merge them with text tokens, prioritizing the retention of text-based KV pairs.

\section{Method}
\begin{figure*}[thbp]
\centering
\includegraphics[width=0.98\linewidth]{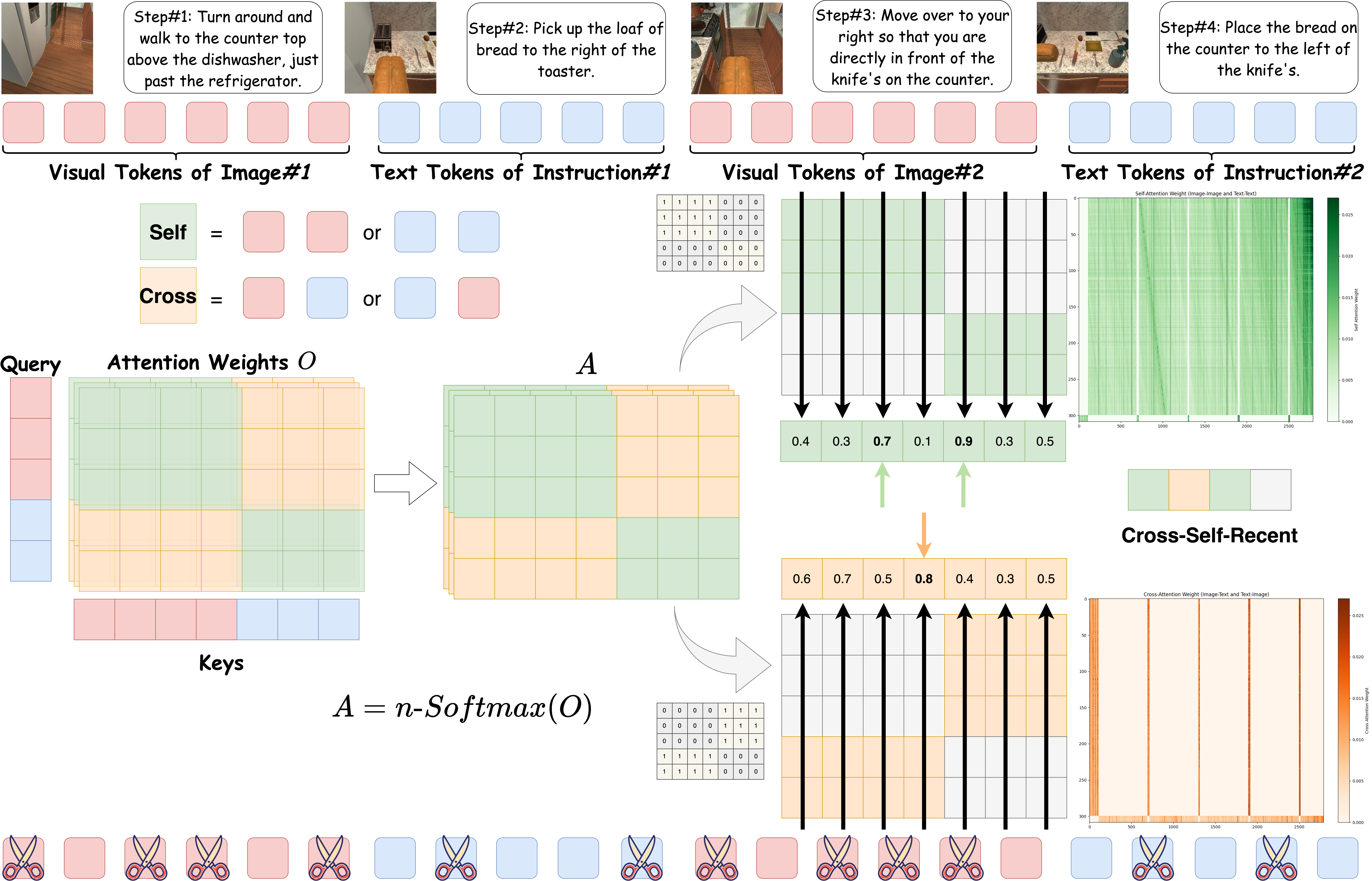} 
\caption{This illustration depicts the \textbf{C}ross-\textbf{S}elf \textbf{P}runing (CSP) KV cache process. The input sequence \(\{ \text{\#image, \#text,\#image, \#text, ...} \}\) is projected onto query and key representations across multiple modalities. The $n$-softmax attention weights serve as the selection function, which is decomposed into intra- and cross-modality. Summation is performed along the query axis within each region, and top-$k$ keys are selected along the key dimension to retain tokens for pruning. 
}
\label{fig:method}
\end{figure*}

Our approach decomposes the attention scores into intra-modality and inter-modality attention. We apply top-k selection within each region, ranking the tokens by summing attention scores along the query dimension for each key. Finally, we concatenate the selected tokens with recent tokens to form the key and value cache.

\subsection{Preliminary}

The auto-regressive generation of text yields a multi-step generation process. At each step $i$, the current token $x_i$ is predicted by the LLM based on the input of prompt and previously-generated tokens $\{x_j\}_{j=1}^{i-1}$.

For reducing the computational cost and avoiding duplicated computations, current inference usually adopts KV cache technique, where the keys ($K_j$) and values ($V_j$) in self-attention of each previous tokens $x_j$ are cached in memory and reused in subsequent steps.

However, when the context length is long, the storage of KV values still poses significant challenges in memory size and memory access speed. Therefore, to reduce the memory cost and run LLMs on resource-limited devices, KV cache pruning methods\cite{li2024snapkv, zhang2024h2o, ren2024efficacy, adnan2024keyformer, liu2024minicache, cai2024pyramidkv, ge2023model} are proposed to remove the less-important tokens.

Generally, the method contains two main components: (1) an importance estimation function $f$ to measure the importance of each token $x_j$; (2) a selecting strategy to keep important tokens in KV and remove the rest based on their importance. Formally, with KV sequences
to be cached, and $T$ denotes the maximum cache length, it first measures the importance scores of each token using $f$, then generate
the selective mask $M, M_i \in \{0, 1\}, \forall i=\{1,\dots,T\}$ to select the tokens with top-T highest scores, where the tokens with zero values in $M$ are omitted.

\subsection{Cross-self pruning}

Previous KV cache pruning methods \cite{li2024snapkv,  ren2024efficacy, zhang2024h2o, liu2024minicache, cai2024pyramidkv, wan2024look} usually use the self-attention scores to indicate the importance of each token, as the attention scores determine the contributions of tokens to the attention output. Nevertheless, some recent works \cite{zhang2024sparsevlm, wan2024look} simply adopt the same strategy for pruning vision-language hybrid tokens, neglecting the distribution gap between different modalities. As validated in Figure~\ref{fig:1}, the discrepancy between attention scores within the same modality and different modalities are significant, resulting in overestimate or underestimate of the importance when considering both modalities together\footnote{Specifically, we find that on VLMs such as LLaVA, the attention scores of text tokens are usually larger than that of visual tokens, potentially leading to the loss of important visual information after KV pruning.}.

Motivated by this, for a more accurate estimation for VLMs, we aim to decompose the attention scores into two parts: intra-modality attention and inter-modality attention. The intra-modality attention denotes the attention scores between tokens within the same modality itself, and inter-modality attention is the ones across different modalities.

Mathematically, for an attention matrix $A\in[0,1]^{L\times L}$ (we average over the head axis if there are multiple heads in attention), where $L=L_t+L_v$ is the text-visual sequence length, we denote $A^{st}\in[0,1]^{L_t\times L_t}$ and $A^{sv}\in[0,1]^{L_v\times L_v}$ as the self-attention scores between text tokens and visual tokens, $A^{ct}\in[0,1]^{L_v\times L_t}$ and $A^{(cv)}\in[0,1]^{L_t\times L_v}$ as the visual-text (text as key) and text-visual cross-attention scores, respectively. Then, we can sum over the query axis to indicate the intra and inter importance of all the keys:
\begin{equation}
\label{eq:cross-self}
    A^{s} = \sum_{k=1}^{L_t}A^{st}_k \oplus \sum_{k=1}^{L_v}A^{sv}_k, \quad
    A^{c} = \sum_{k=1}^{L_t}A^{ct}_k \oplus \sum_{k=1}^{L_v}A^{cv}_k,
\end{equation}
where $\oplus$ denotes concatenation. Then the selective masks $M^{(s)}$ and $M^{(c)}$ of each matrix $A^*$ are obtained by the top-K of the scores:
\begin{equation}
\label{tab:topk}
\scalebox{0.93}{
$\displaystyle
\begin{aligned}
\begin{split}
    M^* = \{M_i\}_{i=1}^{L},
    \quad\mathrm{with\ } M_i^*=\begin{cases} 0, & i\notin\operatorname{top-K}(A^*)\\ 1, & i\in\operatorname{top-K}(A^*)\end{cases}.
\end{split}
\end{aligned}
$
}
\end{equation}

We set different K values $K^{s}$ and $K^{v}$ for intra attention and inter attention, which we find better and will be discussed in experiment section.

Finally, the mask for selecting tokens is from the tokens both important and selected by $M^{s}$ and $M^{c}$:
\begin{equation}
    M = M^{s}\land M^{c}.
\end{equation}

Also note that for better efficiency and accuracy, we trim the attention matrix by a number $O$ of the most recent query tokens as an observation window and a number $R$ of the previous key tokens, \textit{i.e.}, $A[-O:,:-R]$, to reflect the actual needs in the generation of recent context. 
Figure \ref{fig:1} presents the cross-self selection approach. The green part refers to intra-modality, and the orange part to cross-modality. For both regions, we sum along the query axis, rank the score indices, and treat these as the selected key and value indices to retain. Due to potential overlap between the selected candidates \(M^s\) and \(M^c\), the cache budget is significantly optimized across most modality scenarios, resulting in substantial savings in KV cache space. Finally, the combination of cross-self pruned tokens with recent tokens constitutes the optimized KV cache tokens:
\begin{equation}
    \begin{aligned}
        K &= (K \odot M) \oplus K[-R:]\\
        V &= (V \odot M) \oplus V[-R:],
    \end{aligned}
\end{equation}
where $\odot$ is element wise operation.
The pruned key and value are stored in the cache for later inference decoding. Algorithm \ref{alg:csp} presents the cross-self decoding procedure during inference.

\subsection{Smoothness recovery of attention scores}

By using KV cache pruning, we can obtain reduced KV sequences for smaller costs. However, we find there is a sharpness-shift issue in the new pruned attention scores. Let us first consider the original computation of attention scores:
\begin{equation}
    A = softmax(O), \quad \mathrm{with\ } O=(\frac{Q K^T}{\sqrt{d}}),
\end{equation}
where $d$ is a factor for stabilizing the values.

Comparing the attention score $A_i$ between original and post-pruning ones, the difference occurs in the denominator of $softmax$, \textit{i.e.},
\begin{equation} \label{eq:softmax_change}
    \frac{e^{(O_i)}}{\sum_{j\in I^+} e^{(O_j)} + \sum_{j\in I^-} e^{(O_j)}} \rightarrow \frac{e^{(O_i)}}{\sum_{j\in I^+} e^{(O_j )}},
\end{equation}
where $I^+$ and $I^-$ denote the indices of tokens to be kept and pruned, respectively. It is clear to see that the $A_i$ before pruning (left )is smaller than the one after pruning (right), indicating that the original $A$ is smoother. 
These changes would affect the attention outputs by overly enlarging the contributions of tokens with high attention scores, and therefore weaken the performance.

To address this issue, we propose a simple and effective function to recover the original smoothness of attention scores, which we call \textit{n-softmax}, and $A_i$ becomes:
\begin{equation} \label{eq:n-softmax}
    A_i = \textit{n-softmax}(O_i) = \frac{e^{(O_i)}}{n + \sum_{j\in I^+} e^{(O_j )}},
\end{equation}
where $n$ is a hyper-parameter to control the smoothness of the distribution, we set $n=1$ in all experiments.

\begin{algorithm}[!t]
    \caption{\textbf{C}ross-\textbf{S}elf \textbf{P}runing (CSP) Procedure. 
    }
    \label{alg:csp}
    \begin{algorithmic}[1]
        \small
        \State \textbf{Input:} $O \in \mathbb{R}^{H \times L_q \times L_k}$, current keys and values in the cache $K, Q$, budget $T$, recent size $R$, observation window $O$
        \For{iteration $i \gets 1$ to $N$}  
            \State $L_k \leftarrow K$ \Comment{Get key sequence length.}
            \If{$L_k<T$}: \State \textbf{Return:} $K, V$
            \EndIf
            \State $A \leftarrow \textit{n-Softmax}(O)$ \Comment{Select function (Eq. \ref{eq:n-softmax}).}
            \State $A^{st}, A^{sv}, A^{ct}, A^{cv} \leftarrow A$ \Comment{Mask of image and text}
            \State $A^s, A^c \leftarrow A^{st}, A^{sv}, A^{ct}, A^{cv}$ \State \Comment{Acquire cross-attention and intra-attention (Eq. \ref{eq:cross-self}).}
            \State $M^s \leftarrow \text{Topk}(A^{s}) $, $M^c \leftarrow \text{Topk}(A^{c})$ 
            \State \Comment{Get Top-K masks from intra- and cross-modality.}
            \State $M \leftarrow  M^{s}\land M^{c}$ \Comment{Tokens that important both from intra- and cross-modality}
            \State  $K = (K \odot M) \oplus K[-R:]$, 
            \State  $V = (V \odot M) \oplus V[-R:]$ 
            \State \Comment{Concatenate pruned tokens and recent tokens.}
             \State \textbf{Return:} $K$, $V$ \Comment{Pruned Keys and Values}
        \EndFor
    \end{algorithmic}
\end{algorithm}
\vspace{-3mm}



\section{Experiments}
\label{Sec:Experiments}
\begin{table*}[th]
\centering
\begin{tabular}{@{}llllllllllll@{}}
\toprule
Method          & T-1 & T-2 & T-3 & T-4 & S-1 & S-2 & S-3 & S-4 & S-5 & NH & IR \\ \midrule
\multicolumn{12}{c}{\textit{LLaVA-v1.5-7b}}                                              \\ \midrule
Full Cache      & 39.8 & 46.0 & 32.2 & 37.8 & 56.9 & 33.3 & 12.6 & 23.4 & 60.5 & 4.7 & 4.3 \\ \midrule
H2O  \cite{zhang2024h2o}           & 39.5 & 46.6 & 31.8 & 38.5 & 55.0 & 33.0 & 13.0 & 23.0 & 60.0 & 1.4 & 3.7 \\
SnapKV \cite{li2024snapkv}          & 39.6 & 46.0 & 31.5 & 40.6 & 54.6 & 33.6 & 13.0 & 20.0 & 60.0 & 1.4 & 3.7 \\
ReCo \cite{ren2024efficacy}           & 39.7 & 46.1 & 31.8 & 38.5 & 55.0 & 33.0 & 12.6 & 22.8 & 60.0 & 4.7 & 4.3 \\

LOOK-M (A-Merge) \cite{wan2024look}         & 39.7 & 46.1 & 32.2 & 39.1 & 54.9 & 34.0 & 12.4 & 21.4 & 60.5 & 1.6 & 3.7 \\
LOOK-M (W-Merge) \cite{wan2024look}        & 39.6 & 46.1 & 31.8 & 39.1 & 55.1 & 34.0 & 13.2 & 24.0 & 60.5 & 1.4 & 3.7 \\ 
LOOK-M (P-Merge) \cite{wan2024look}        & 39.7 & 46.1 & 32.5 & 39.9 & 57.0 & 34.0 & 12.8 & 23.9 & 60.5 & 5.3 & 3.8 \\ 
LOOK-M (TP+A-Merge) \cite{wan2024look}        & 39.7 & 46.1 & 32.0 & 39.0 & 56.5 & 33.8 & 12.9 & 23.1 & 60.5 & 5.1 & 3.5 \\ 
LOOK-M (TP+W-Merge)  \cite{wan2024look}       & 39.8 & 46.1 & 32.5 & 39.9 & 57.0 & 34.0 & 12.8 & 23.9 & 60.5 & 5.3 & 3.8 \\ 
LOOK-M (TP+P-Merge) \cite{wan2024look}        & 39.8 & 46.1 & 32.5 & 39.9 & 57.0 & 34.0 & 12.8 & 23.9 & 60.5 & 5.3 & 3.8 \\ 
\midrule
CSP (Ours) &  \textbf{39.9}   &  \textbf{46.8}  &  \textbf{32.5}   &  \textbf{41.6}   &  \textbf{57.5}   &  \textbf{34.1}   &  \textbf{13.7}   &  \textbf{27.8}   &  \textbf{61.0}   & 1.4   &  \textbf{6.3}  \\ \midrule 
\multicolumn{12}{c}{\textit{LLaVA-v1.5-13b}}                                              \\ \midrule
Full Cache      & 40.0 & 46.0 & 32.2 & 37.8 & 56.9 & 33.3 & 12.6 & 23.4 & 60.5 & 4.7 & 4.3 \\ \midrule
H2O \cite{zhang2024h2o} & 39.5 & 45.9 & 30.4 & 47.9 & 64.1 & 48.7 & 13.9 & 25.1 & 59.7 & 3.6 & 0.0 \\
SnapKV \cite{li2024snapkv} & 39.6 & 46.0 & 30.6 & 47.8 & 64.2 & 48.2 & 13.4 & 22.9 & 59.8 & 4.2 & 1.0 \\
ReCo \cite{ren2024efficacy} & 39.7 & 45.9 & 30.5 & 48.0 & 64.3 & 48.3 & 13.8 & 24.9 & 59.7 & 3.5 & 0.0 \\ 
LOOK-M (A-Merge) \cite{wan2024look} & 39.7 & 46.1 & 30.7 & 48.0 & 64.6 & 48.0 & 13.3 & 22.1 & 59.8 & 4.6 & 1.0 \\
LOOK-M (W-Merge) \cite{wan2024look} & 39.6 & 46.1 & 30.6 & 47.9 & 64.5 & 48.4 & 13.4 & 23.4 & 59.9 & 4.7 & 1.0 \\
LOOK-M (P-Merge) \cite{wan2024look} & 39.7 & 46.0 & 30.6 & 48.0 & 64.6 & 48.0 & 13.3 & 25.7 & 59.8 & 5.8 & 1.0 \\
LOOK-M (TP + A-Merge) \cite{wan2024look} & 39.7 & 46.2 & 30.7 & 48.0 & 65.4 & 48.3 & 13.7 & 26.6 & 60.0 & 11.2 & 1.0 \\
LOOK-M (TP + W-Merge) \cite{wan2024look} & 39.8 & 46.1 & 30.8 & 48.1 & 64.8 & 48.2 & 13.9 & 26.9 & 60.0 & 11.4 & 1.0 \\
LOOK-M (TP + P-Merge) \cite{wan2024look} & 39.8 & 46.2 & 30.8 & 48.1 & 65.2 & 48.5 & 14.1 & 26.6 & 60.0 & 11.7 & 1.0 \\ \midrule
CSP  (Ours)        &   \textbf{40.0}  &  \textbf{46.5}   &  \textbf{32.4}   & \textbf{49.0}    &  \textbf{65.4}  &  48.3  &  \textbf{14.4}   &  \textbf{27.0}   &  \textbf{60.5}   &  3.1  &  \textbf{9.6}  \\ \bottomrule
\end{tabular}
\caption{Comparison of various KV cache management methods on the multiple multimodal tasks. Tasks are grouped as follows: Temporal Multi-Image Tasks (T1-T4), Semantic Multi-Image Tasks (S1-S4), Needle in a Haystack (NH), and Image Retrieval (IR). For fair comparison, we set the same KV cache size for all cache methods. Scores are calculated as the average performance across datasets within each subtask. For fair comparison, we set the same KV cache size for all inference methods.} \label{tab:tab1}
\vspace{-3mm}
\end{table*}

\subsection{Benchmark}
We evaluate our method on MileBench \cite{song2024milebench} (MLLM), a benchmark specifically designed to assess long-context capabilities in multimodal language models. It collects widely-used datasets, providing a versatile and realistic foundation for evaluating model inference performance through two evaluation sets, \textit{diagnostic} and \textit{realistic-crafted}, which systematically measure inference performance in multi-modality scenarios. The tasks in the benchmark organized with:
\begin{itemize}
    \item \textbf{Temporal Multi-Image Tasks (T1-T4)}: Temporal tasks involve understanding and predicting sequential events across images, and the methods in handling in action recognition, object tracking and spatial navigation.
    \item \textbf{Semantic Multi-Image Tasks (S1-S4)}: Semantic tasks focus on interpreting multimodal information, requiring inference methods to reason the knowledge-based QA, text-rich image analysis, visual relationship inference, dialogue understanding, and spatial reasoning.
    \item \textbf{Needle in a Haystack (NH)}: Retrieval tasks designed to find a preset password from a long context, which test kv cache inference methods in precise password extraction.
    \item \textbf{Image Retrieval (IR)}: Focused on identifying target images from candidates, which assess KV cache methods’ effectiveness in perceptual and conceptual recognition. 
\end{itemize}
Details of these challenging and comprehensive multimodal tasks, which include the corresponding datasets and evaluation metrics, are provided in the Appendix.

\subsection{Baselines}
We compare our kv cache method with previous mainstream baselines include SnapKV \cite{li2024snapkv}, H2O \cite{zhang2024h2o}, ReCo \cite{ren2024efficacy}, and LOOK-M \cite{wan2024look}, each offering unique strategies for managing KV cache in long-context scenarios. SnapKV \cite{li2024snapkv} introduces a method for intelligently identifying attention allocation patterns and compressing the KV cache by pooling essential tokens for extended sequences. H2O \cite{zhang2024h2o} presents a KV cache eviction policy that dynamically balances recent and frequently accessed tokens, identifying "heavy hitters" solely based on attention scores. ReCo \cite{ren2024efficacy} focuses on enhancing the efficacy of existing eviction policies through refined importance score calculations and carefully constructed eviction scopes, proposing a robust cache omission policy rooted in temporal attention scores and robustness measures. The LOOK-M family \cite{wan2024look} considers modality awareness to evict image tokens and merge them with text tokens. This method prioritizes the retention of text-based KV pairs while evicting image-based KV pairs. The merging strategies include Average Merging (A), which computes the mean value of tokens within the similarity matrix for merging; Weighted Merging (W), which dynamically adjusts token weights based on similarity scores for adaptive merging; and Pivotal Merging (P), which enhances the importance of key conserved tokens during the merging process.

\subsection{Setup}
We employed LLaVA-v1.5-7b \cite{liu2024improved} on RTX-4090 GPUs with \textit{flash-attn-2.4.3post1} and LLaVA-v1.5-13b \cite{liu2024improved} on A100 GPUs with \textit{flash-attn-2.6.3} \footnote{https://github.com/Dao-AILab/flash-attention} to conduct our experiments. To maintain consistency in generation, we set the sampling method to deterministic with a fixed temperature of 0, and the maximum context length was configured to 4096 tokens. Batch sizes were dynamically set based on dataset characteristics to balance computational load and memory constraints. Specifically, for datasets MMCoQA \cite{li2022mmcoqa}, NeedleInAHaystack \cite{song2024milebench}, and GPR1200 \cite{schall2022gpr1200}, the batch size was set to 1, while for all other datasets, a batch size of 24 was employed. Additionally, we calibrated the top-k value selection ratio for self-attention and cross-attention based on the sample mean ratio of cross-self region, with ablation studies showing the efficacy of adjusting this ratio. We applied biases of $0.5$ and 1.5 in EgocentricNavigation\cite{krantz2020beyond} and SlideVQA \cite{tanaka2023slidevqa}, respectively, while keeping the default settings for other datasets. Our pre-processing and evaluation pipeline follows the standards of the benchmark, ensuring consistent assessment across the widely-used 29 datasets. 




\subsection{Main results}
We use the widely-adopted open-source vision-language model LLaVA \cite{liu2024improved} to test KV cache performance on the benchmark. We present the results of our experiments in Table \ref{tab:tab1} and summarize our findings as follows.

For the LLaVA-v1.5-7b model, our approach achieves notable improvements across several tasks. By independently selecting top-k tokens for cross-attention and self-attention, our method effectively retains key tokens specific to each modality. This separation enables the model to capture essential temporal sequences in tasks with sequential dependencies while simultaneously focusing on relevant multimodal content in tasks requiring semantic understanding. This result reveals that separating cross and self attentions allows for better retention of modality-specific cues, enhancing performance in tasks highly relevant to visual and textual data, with improvements of 4.5\%, 7.2\%, and 9.8\% in T-3, S-5, and 4.5\%, 7.2\%, 9.8\% improvement in NH task respectively.

For the larger model, LLaVA-v1.5-13b, our approach shows even more pronounced improvements, especially on tasks T3, T4, and IR. These tasks share a common demand for precise handling of spatial and sequential elements across visual and textual modalities. By separating cross-attention and self-attention during top-k selection, our method effectively retains modality-specific tokens, which is crucial for tasks requiring spatial alignment and temporal tracking. This selective retention allows the model to preserve essential visual details for spatial localization (T3) with a performance boost of 8.3\%, state transition (T4) with a 7.2\% increase, and accurate retrieval in IR with a 9.6\% improvement.

\section{Ablation Study}
\label{sec:ablation}

We delve into ablation analysis of the KV cache approach to comprehensively assess our method. First, we present the hyper-parameters selection of $K^c$ and $K^s$. Next, we assess speed latency and GPU memory usage to examine efficiency. We also test the pruning selection function, and record the impact of varying budget sizes on performance. Finally, we present the influence of model architectures by introducing other vision-language models.

\subsection{Influence of the hyper-parameters}
The remaining tokens in the cache are composed of the top-$k$ indices selected independently from self-attention and cross-attention, which are then concatenated with recent tokens. Consequently, the hyperparameters of our method include the ratio of top-$k$ selections between cross-attention and self-attention.
\begin{figure}
  \centering
  \begin{subfigure}{0.48\linewidth}
    \includegraphics[width=\textwidth]{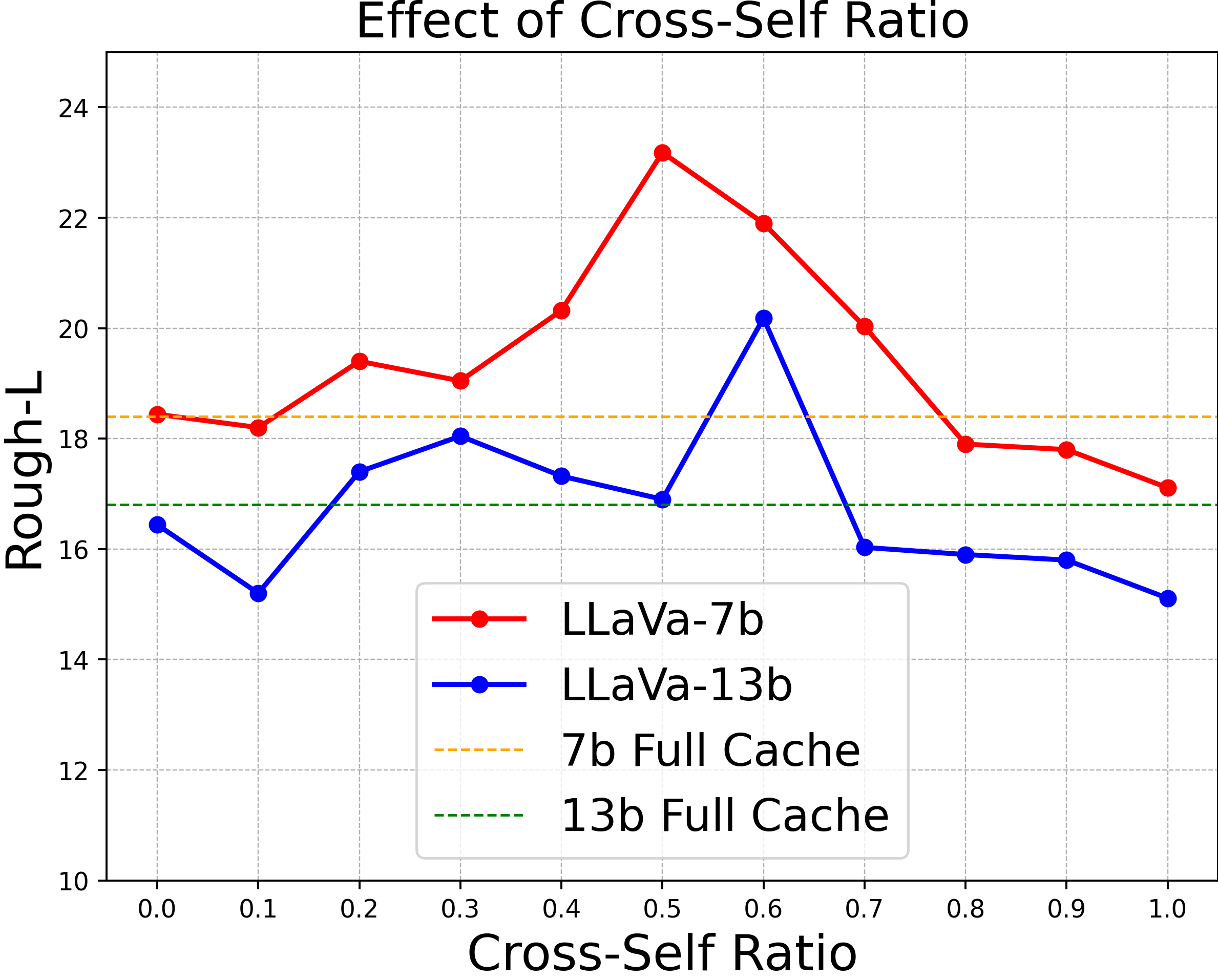} 
    \caption{ALFRED.}
    \label{fig:A1_1}
  \end{subfigure}
  \hfill
  \begin{subfigure}{0.48\linewidth}
    \includegraphics[width=\textwidth]{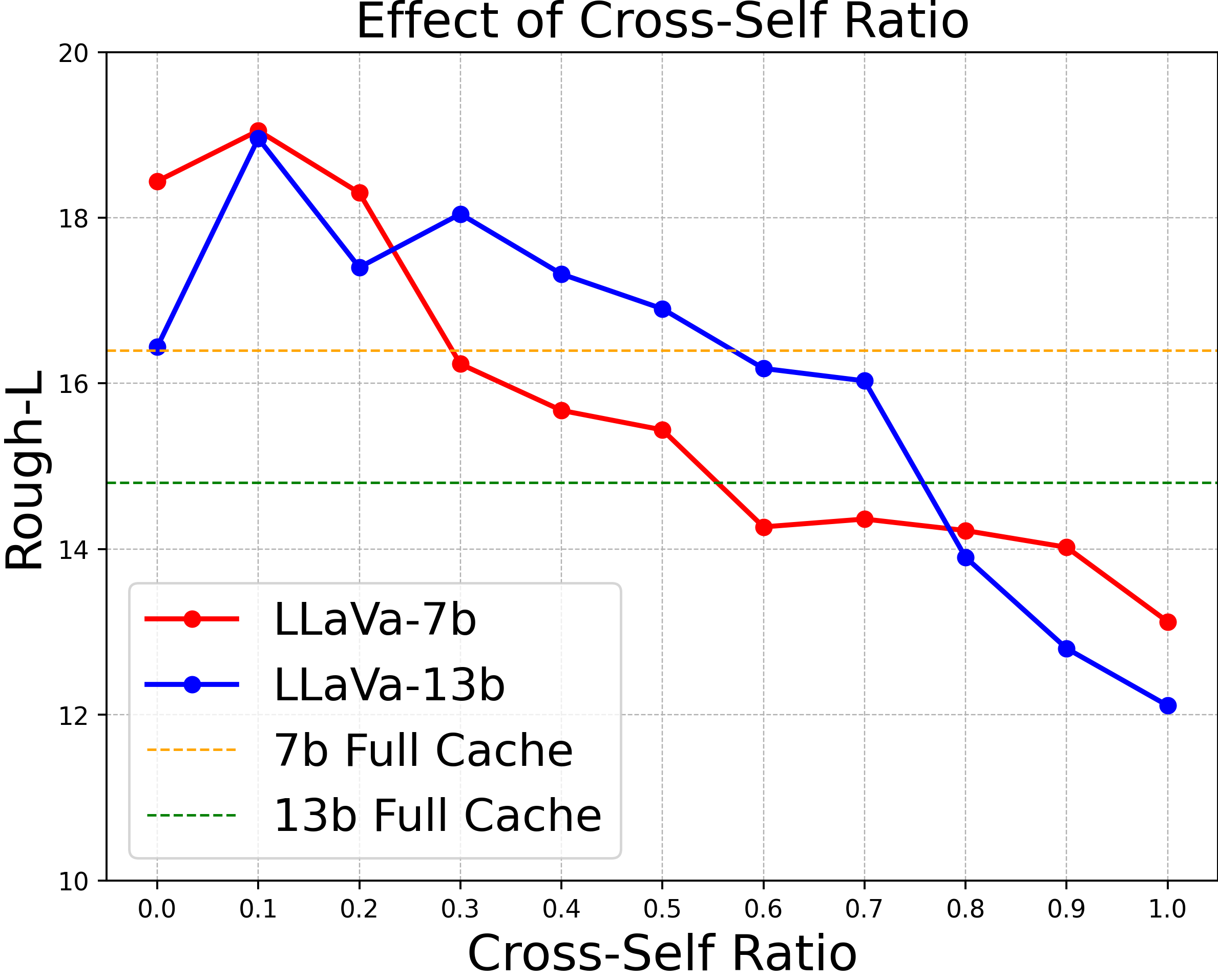}
    \caption{Clever-Change}
    \label{fig:A1_2}
  \end{subfigure}
  \hfill
  \begin{subfigure}{0.48\linewidth}
    \includegraphics[width=\textwidth]{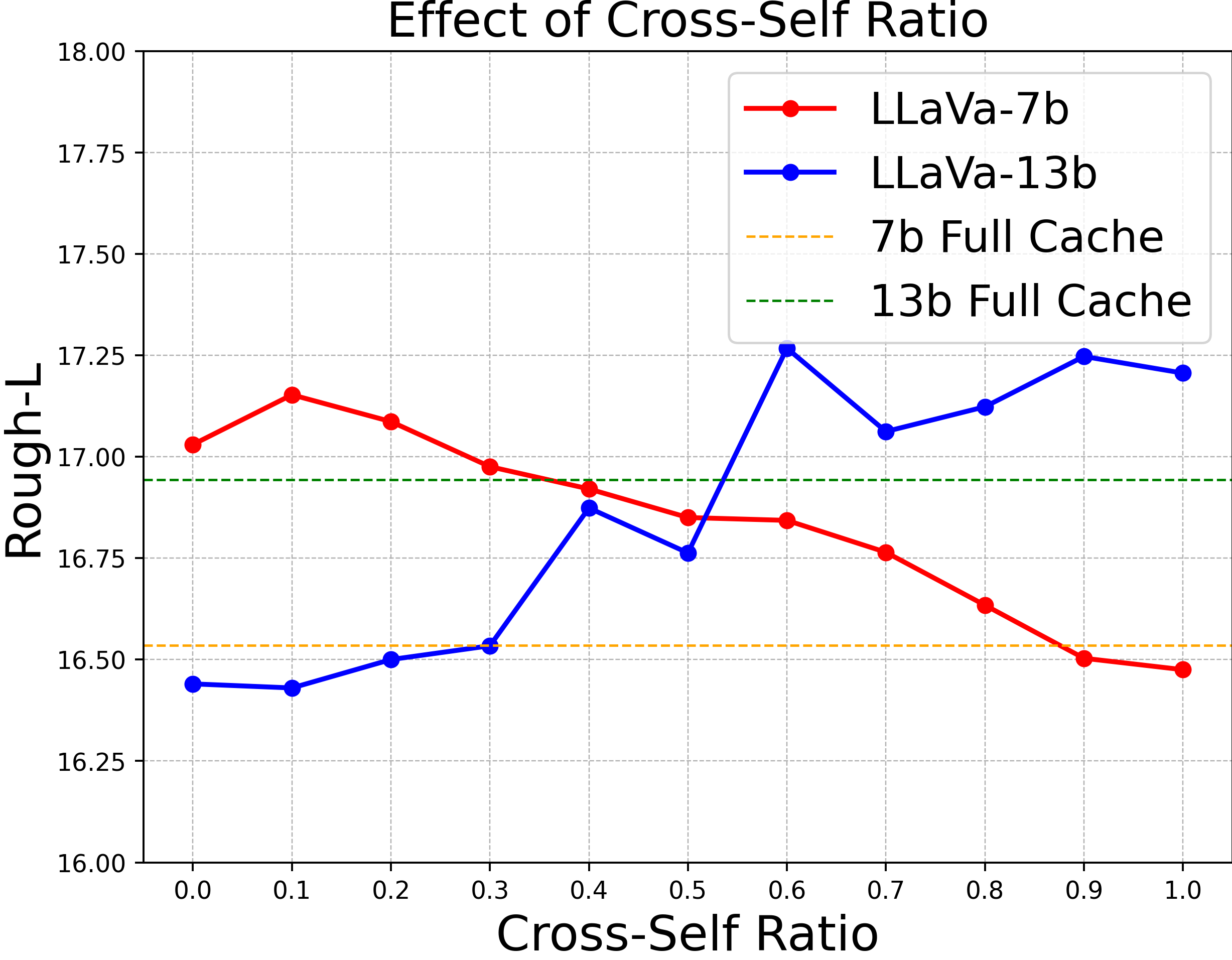}
    \caption{Spot-the-Diff}
    \label{fig:A1_3}
  \end{subfigure}
  \hfill
  \begin{subfigure}{0.48\linewidth}
    \includegraphics[width=\textwidth]{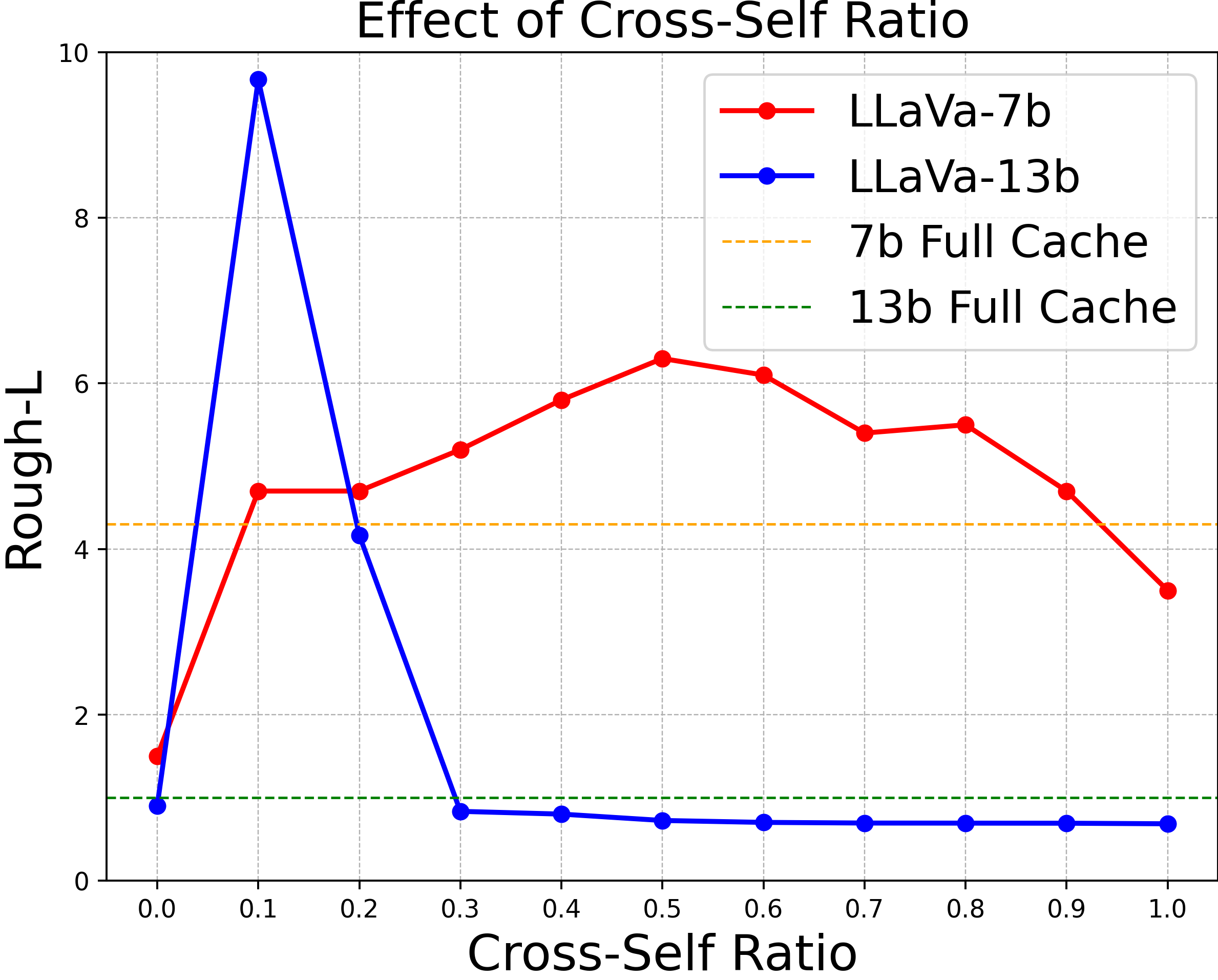}
    \caption{GPR2000}
    \label{fig:A1_4}
  \end{subfigure}
  \vspace{-2mm}
  \caption{The impact of the cross-self ratio.}
  \label{fig:A1}
  \vspace{-6mm}
\end{figure}
We observe that performance reaches an optimal level when both cross-attention and self-attention tokens are selected in balanced proportions (i.e., the ratio is neither 0 nor 1) across 29 datasets. This configuration suggests that integrating both types of attention improves performance across datasets, as it allows the model to capture important tokens from multiple perspectives. However, there are differences between datasets; for instance, tasks focused on temporal alignment or sequential event detection (e.g. ALFRED), tend to benefit more from cross-attention, where context from multiple image frames is critical. On the other hand, datasets emphasizing individual object identification or attribute-focused reasoning are more reliant on self-attention, as they require maintaining a focused view on specific elements within a single frame.

In extreme cases (ratio = 1 for only self-attention 
$K^s$ and ratio = 0 for only cross-attention 
$K^c$), we find distinct effects on task performance. With self-attention, static tasks like visual recognition perform well due to precise frame-specific representations, though this setup struggles with tasks needing cross-frame integration. Conversely, cross-attention supports tasks requiring sequence alignment, like scene understanding, but can dilute focus on high-resolution details crucial for object-specific tasks.

\subsection{Efficiency analysis}
\label{sec:A2}
We analyze the efficiency of our proposed method in terms of decoding and GPU memory usage. Table \ref{tab:A2} presents the results using LLaVA-v1.5-7b tested on a single RTX-4090 GPU with 100 warm-up iterations.The samples tested for the latency and memory usage is sampled from the ALFRED dataset in the benchmark. 
At a 60\% budget, decoding latency is reduced to 24.377 ms/token, which provides a modest improvement over the 26.023 ms observed with a full cache, while memory usage drops by approximately 23\% to 1.207GiB. As the cache budget decreases further, the benefits become more pronounced: at a 30\% budget, latency reaches 21.027 ms per token and memory usage is nearly halved to 0.523 GiB. With the most aggressive reduction, using only 10\% of the original cache, decoding latency improves to 16.287 ms per token, achieving a 37\% speed increase over the full cache setup and an 87\% reduction in memory usage to just 0.208 GiB.
These findings illustrate that our method allows for a flexible balance between memory efficiency and processing speed. Even with significantly reduced cache budgets, our approach retains acceptable latency and memory performance, offering a scalable solution for resource-constrained environments. 

\begin{table}[th]
\centering
\resizebox{\linewidth}{!}{
\begin{tabular}{@{}lccc@{}}
\toprule
Method     & Budget & Decoding Latency & GPU Mem \\ \midrule
Full Cache & 100\%  &    26.023 ms/token             &   1.571 GiB      \\
CSP        & 60\%   &    24.377 ms/token              &   1.207 GiB
      \\
CSP        & 30\%   &    21.027 ms/token              &   0.523 GiB      \\
CSP        & 20\%   &    19.736 ms/token              &   0.476 GiB      \\
CSP        & 10\%   &   16.287 ms/token               &         0.208 GiB \\ \bottomrule
\end{tabular}
}
\caption{The efficiency of latency speed and memory usage. We utilized LLaVA-v1.5-7b to test speed performance on a single RTX-4090 with 100 warm-up iterations.}
\label{tab:A2}
\vspace{-6mm}
\end{table}

\subsection{Influence of n-softmax}
In this ablation study, we compare two KV cache pruning methods: one selects the top-k tokens directly from the overall selection function, while the other applies top-k selection separately within cross-attention and self-attention regions. We evaluate both approaches with standard Softmax and $n$-Softmax scoring functions to assess their impact on performance. Experimental results reveal that $n$-Softmax consistently provides a slight performance improvement over Softmax, indicating that the smooth transition positively impacts decoding performance. Specifically, selection-based approaches demonstrate clear benefits by focusing retention on high-value tokens, which enhances model efficiency. This effect is evident across tasks, as the separate top-k selection in cross and self-attention regions improves performance by capturing modality-specific important tokens more effectively. Furthermore, we find that this transition is particularly effective for tasks requiring both temporal coherence and fine-grained feature retention, as it allows for the selective pruning of large, less critical tokens under limited KV cache budgets. These results underscore the advantages of selection-based methods for KV cache pruning, especially when integrating cross-self separation with $n$-Softmax. 


\begin{figure}
  \centering
  \begin{subfigure}{0.48\linewidth}
    \includegraphics[width=\textwidth]{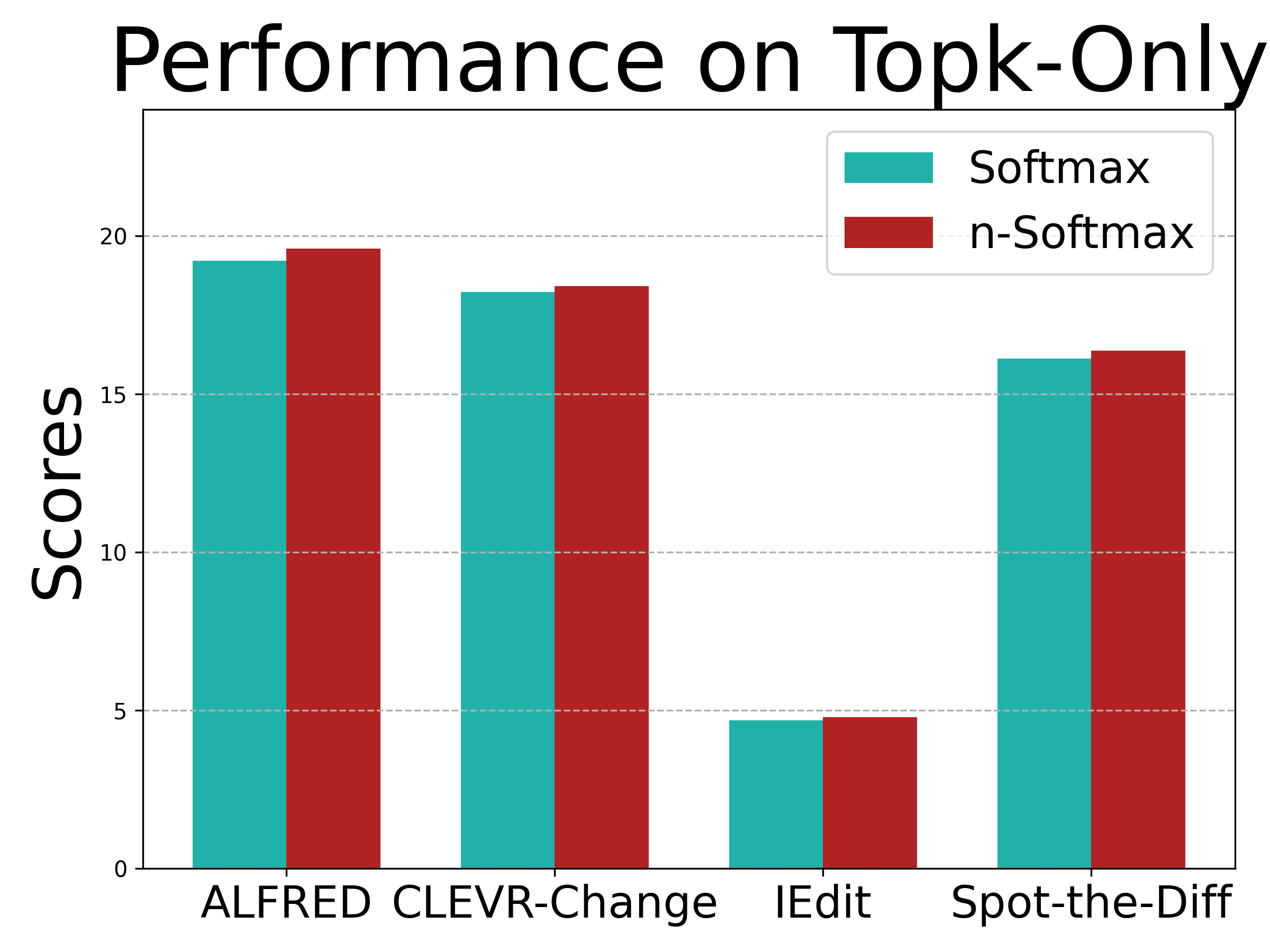} 
    \caption{n-Softmax on Top-k selection.}
    \label{fig:A3_1}
  \end{subfigure}
  \hfill
  \begin{subfigure}{0.48\linewidth}
    \includegraphics[width=\textwidth]{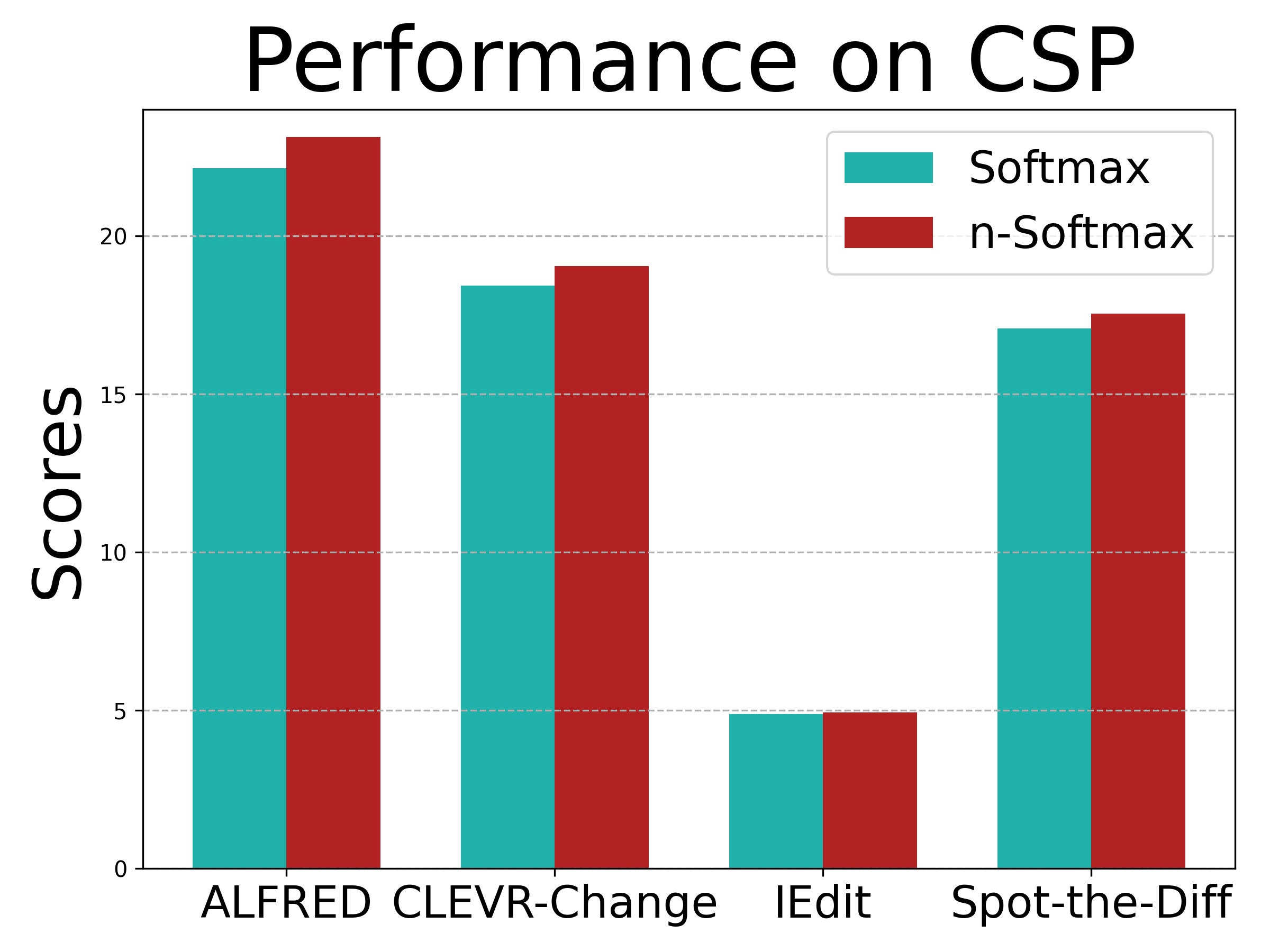} 
    \caption{n-Softmax on cross-self.}
    \label{fig:A3_2}
  \end{subfigure}
  \caption{The benefit of n-softmax. We conduct the experiments on the ALFRED dataset by LLaVA-v1.5-7b.}
  \label{fig:short}
  \vspace{-6mm}
\end{figure}

\subsection{Impact of cache budget}
\label{sec:A4}
As the cache budget increases, we observe consistent performance gains across all tasks, indicating that larger cache budgets enhance model accuracy and retrieval quality. 
For each dataset, performance improves steadily with an increase in cache size, moving closer to or exceeding the baseline set by full cache. In tasks with complex sequence dependencies like ALFRED, our method (CSP) achieves a significant boost in accuracy at higher cache budgets (60\%), outperforming other methods and reaching a level above the full cache baseline. This pattern suggests that a larger cache budget is especially beneficial in scenarios where maintaining temporal coherence is crucial for task success. In tasks requiring fine-grained visual differentiation, such as Spot-the-Diff and CLEVR-Change, performance gains with increased cache are less pronounced but still evident, indicating that these tasks can benefit from a moderate cache size. 
These findings support that our method consistently outperforms other methods across all budget size in the cache, suggesting that separate top-k selections from cross and self regions could effectively balance tokens selection and avoid collapse in the inference process. 
\begin{figure*}[!th]
  \centering
  \begin{subfigure}{0.24\linewidth}
    \includegraphics[width=\textwidth]{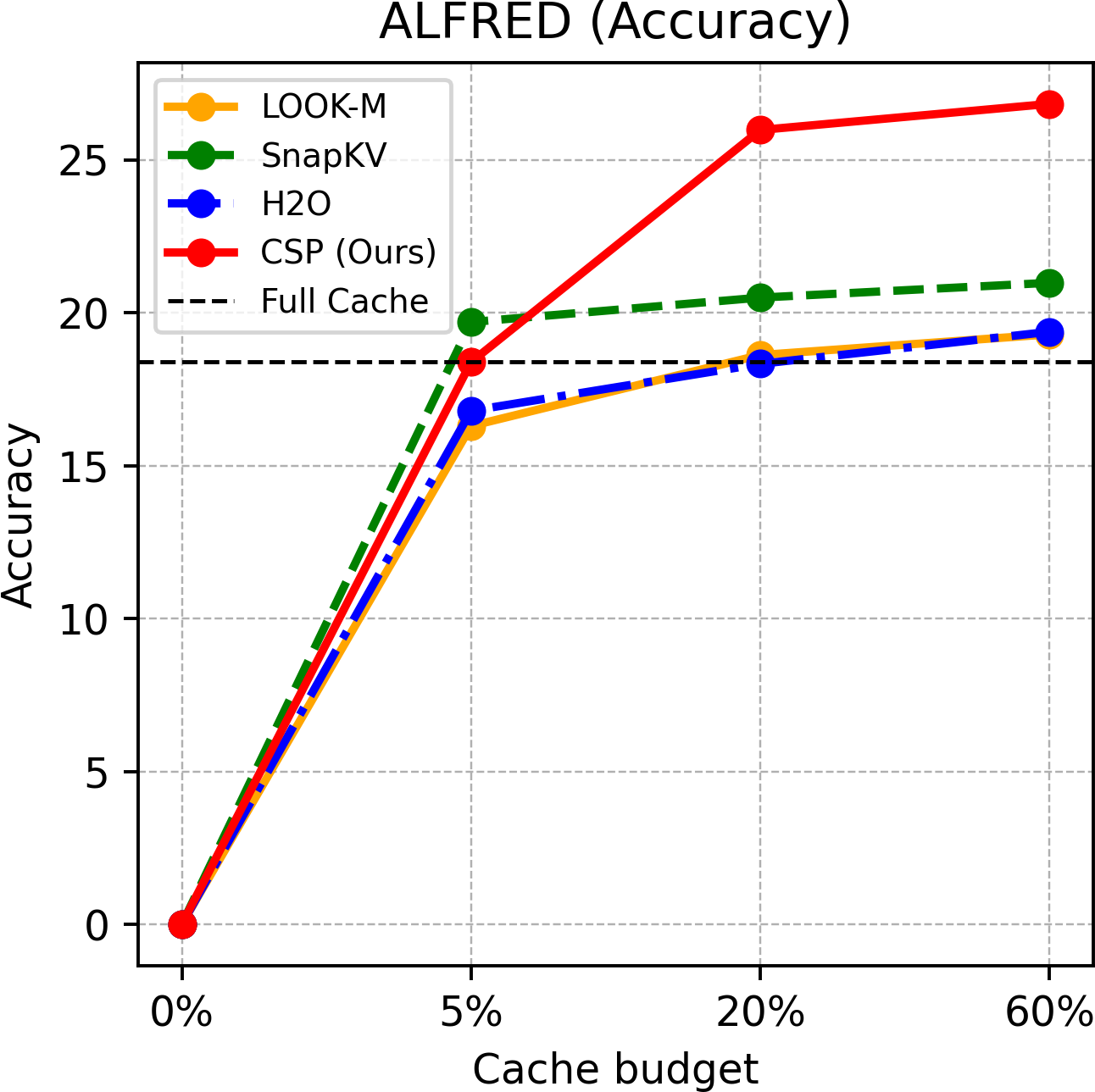} 
    \caption{ALFRED.}
    \label{fig:A4_1}
  \end{subfigure}
  \hfill
  \begin{subfigure}{0.24\linewidth}
    \includegraphics[width=\textwidth]{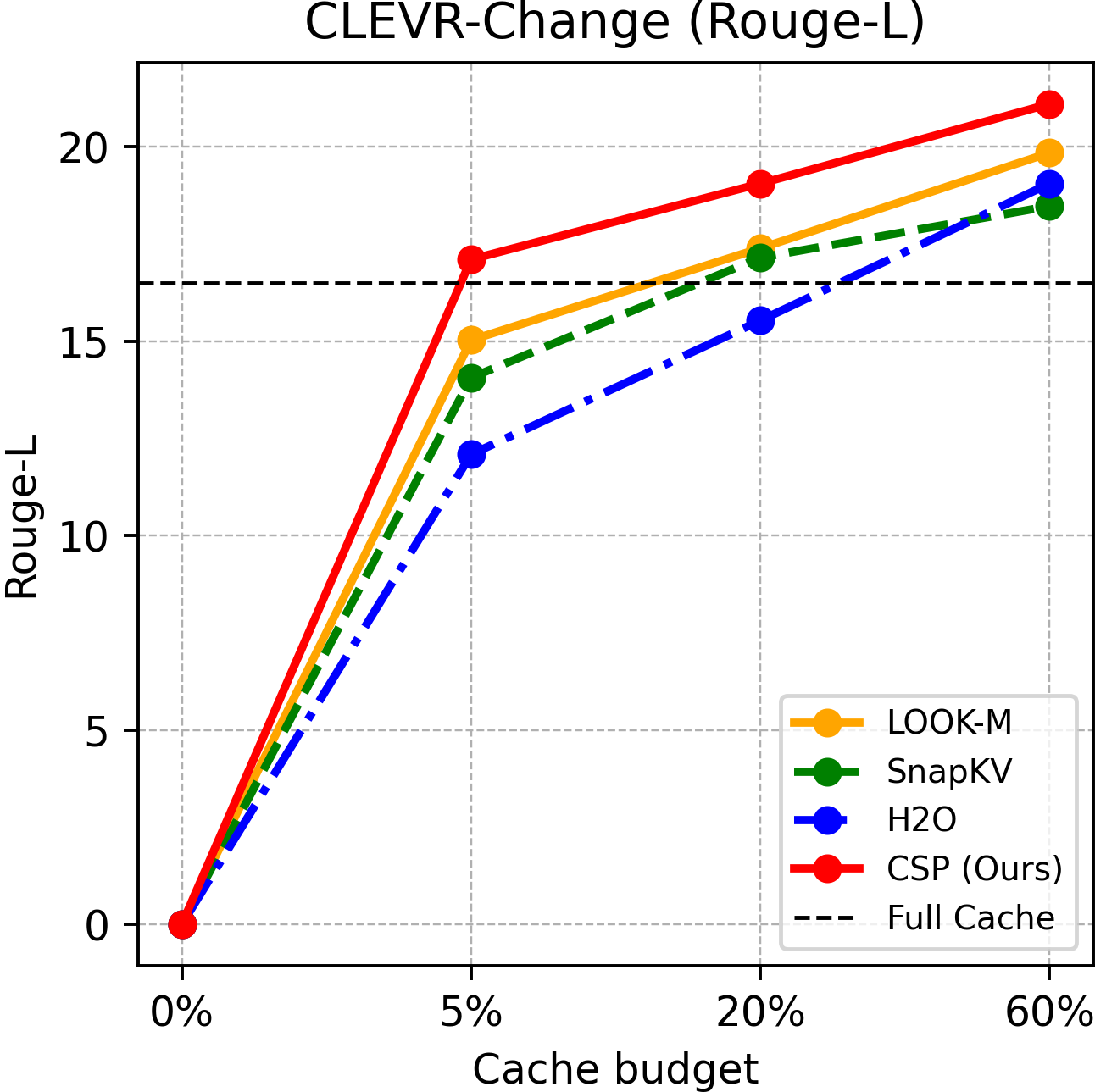}
    \caption{Clever-Change}
    \label{fig:A4_2}
  \end{subfigure}
  \hfill
  \begin{subfigure}{0.24\linewidth}
    \includegraphics[width=\textwidth]{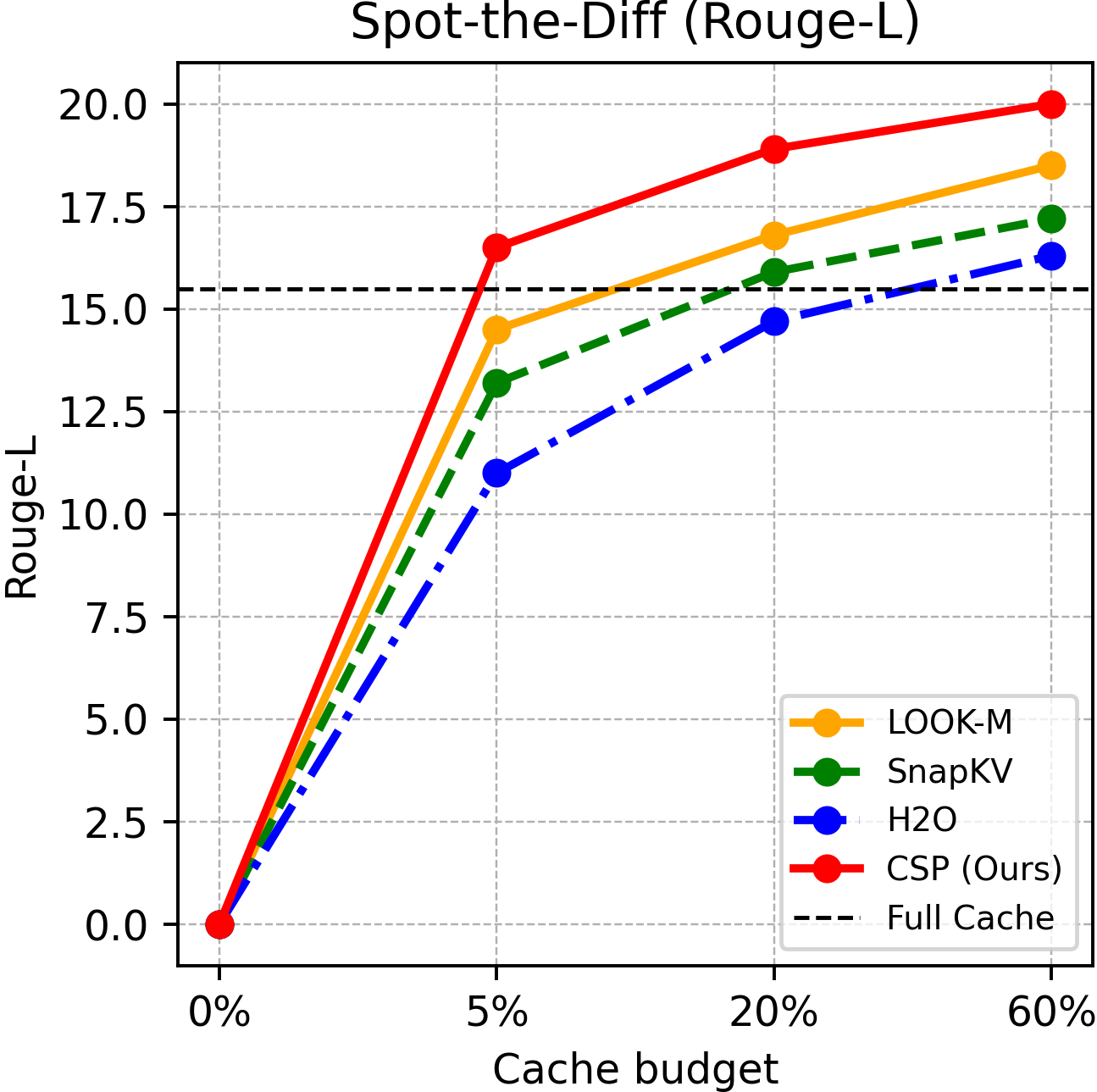}
    \caption{Spot-the-Diff}
    \label{fig:A4_3}
  \end{subfigure}
  \hfill
  \begin{subfigure}{0.24\linewidth}
    \includegraphics[width=\textwidth]{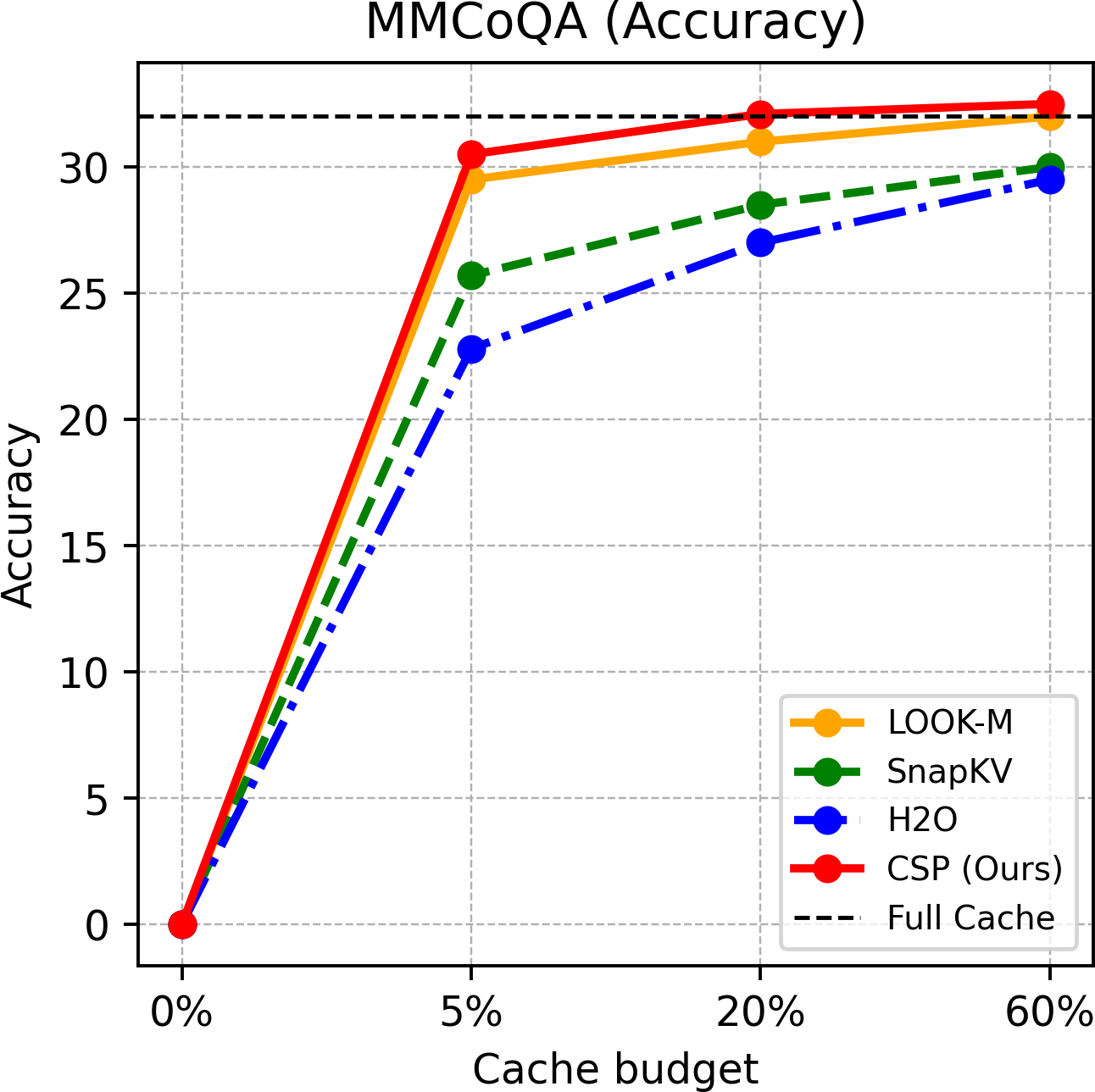}
    \caption{MMCoQA}
    \label{fig:A4_4}
  \end{subfigure}
  \begin{subfigure}{0.24\linewidth}
    \includegraphics[width=\textwidth]{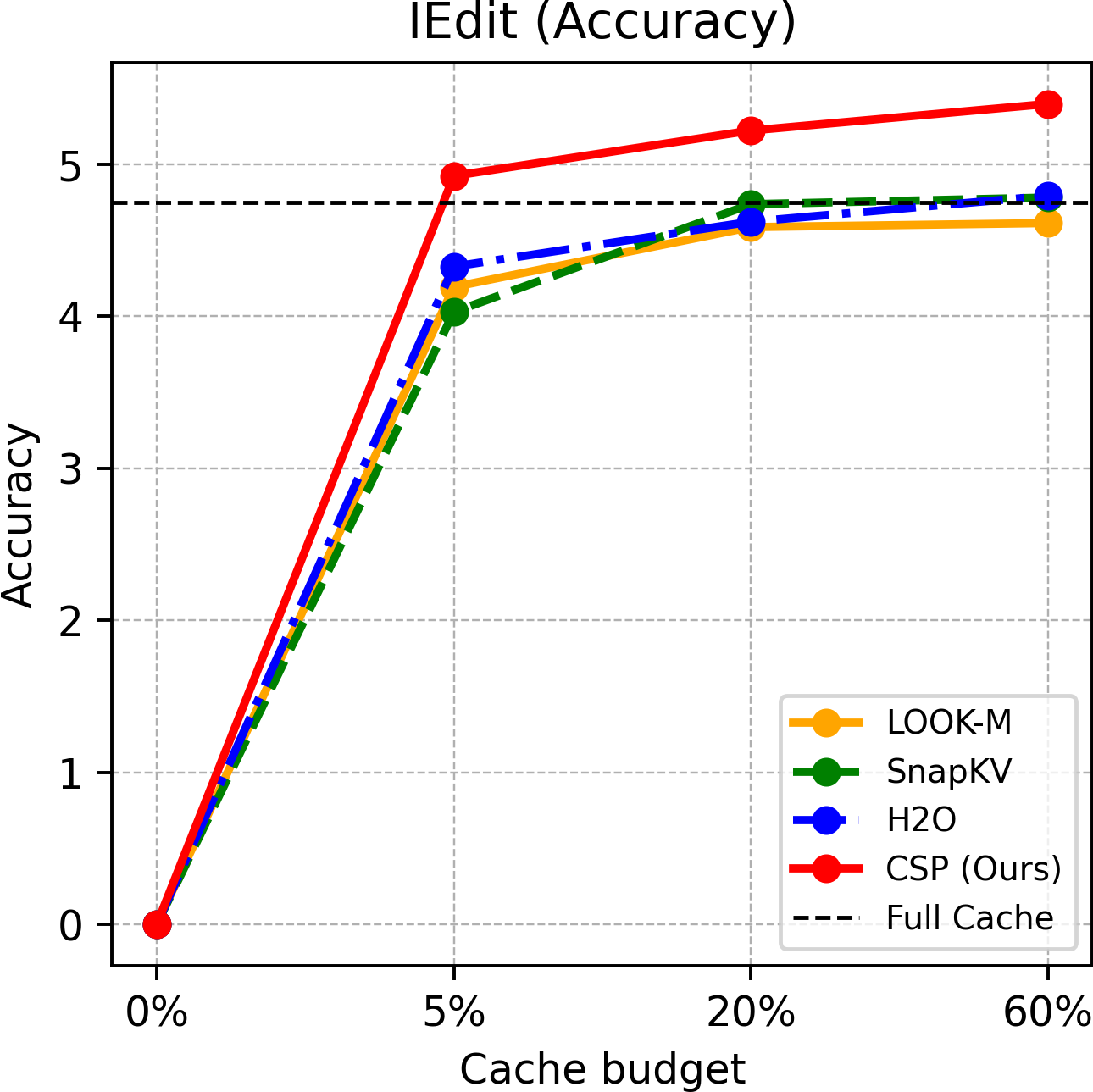} 
    \caption{IEdit}
    \label{fig:A4_5}
  \end{subfigure}
  \hfill
  \begin{subfigure}{0.24\linewidth}
    \includegraphics[width=\textwidth]{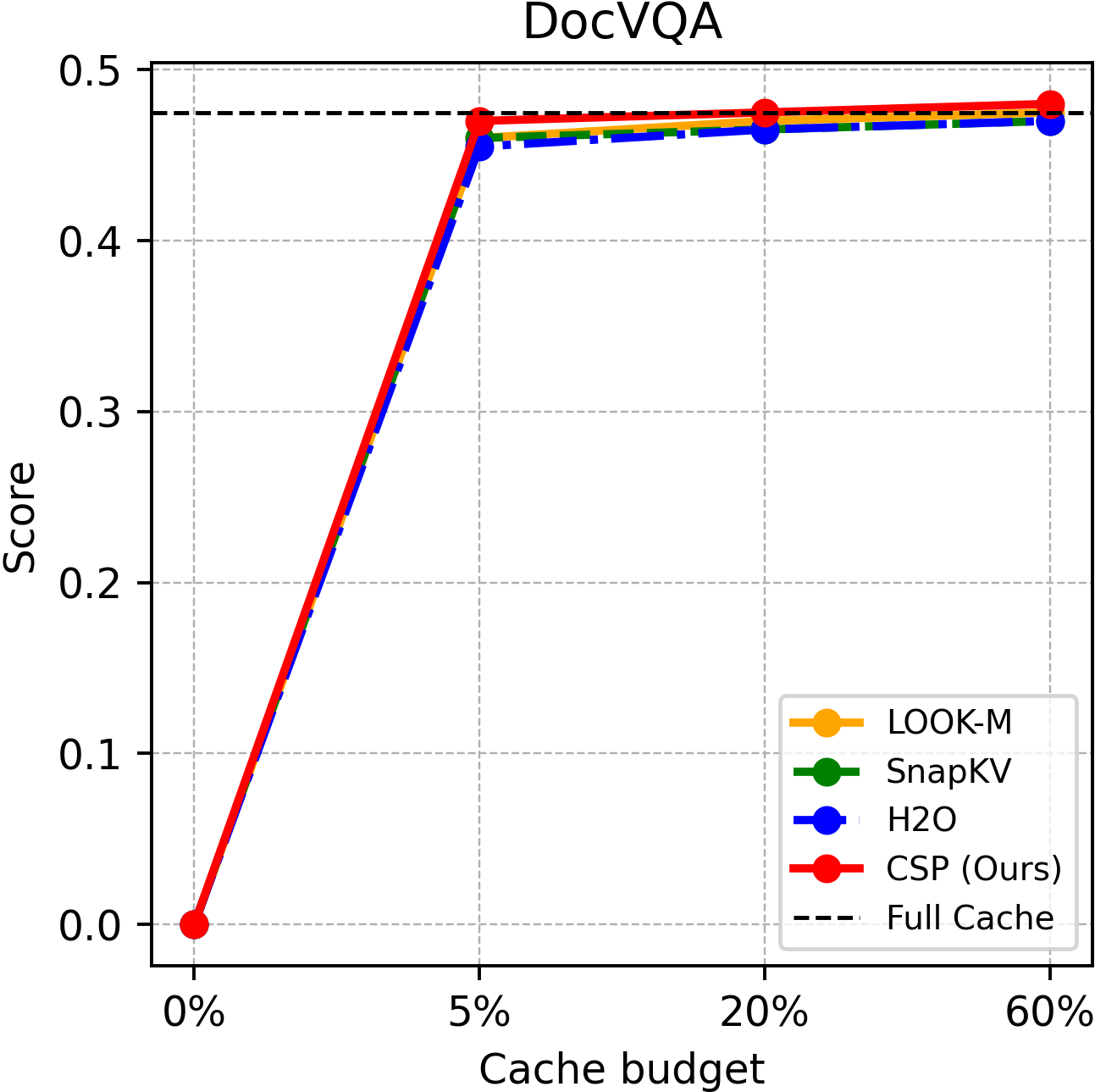}
    \caption{DocVQA}
    \label{fig:A4_6}
  \end{subfigure}
  \hfill
  \begin{subfigure}{0.24\linewidth}
    \includegraphics[width=\textwidth]{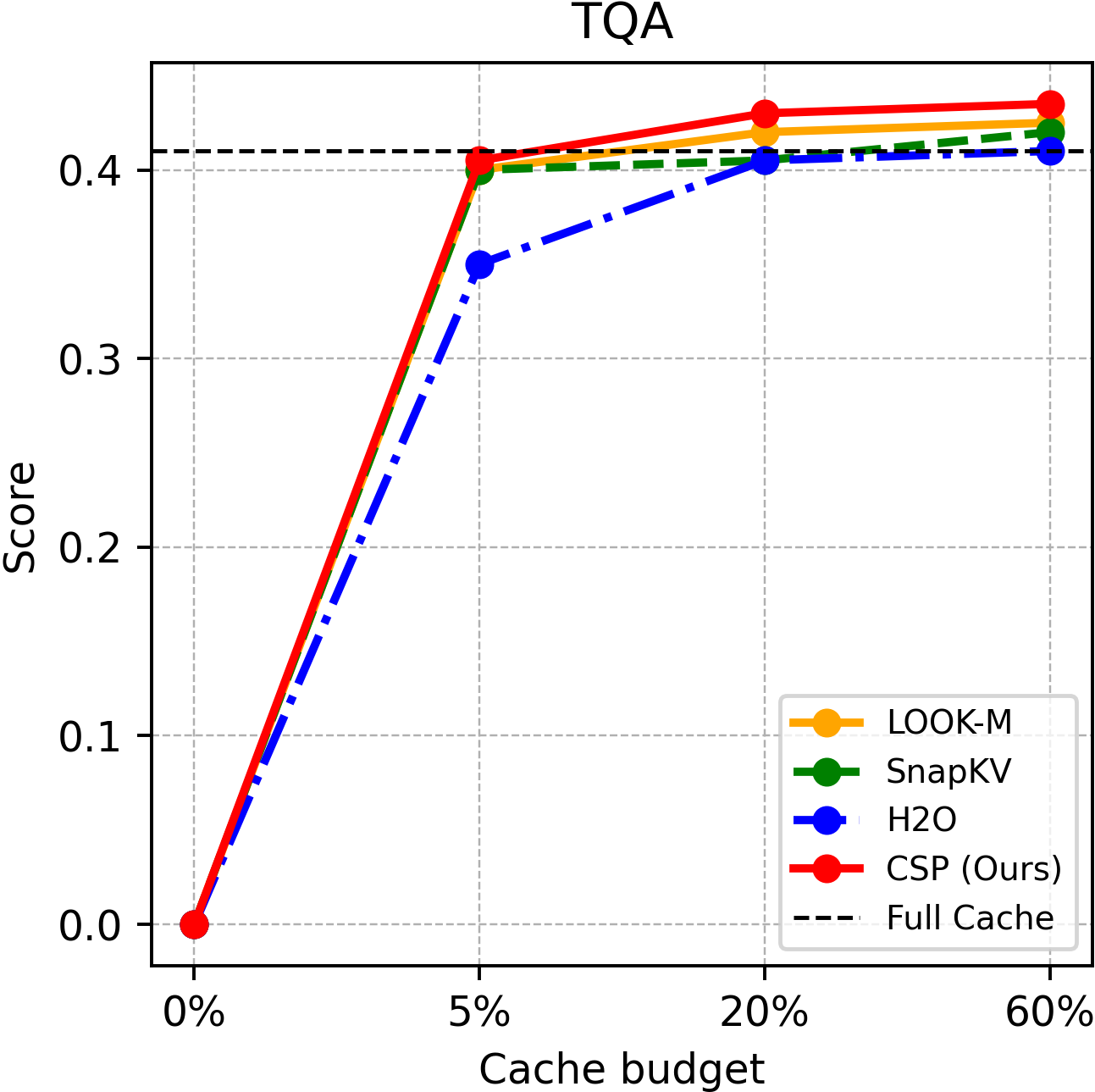}
    \caption{TQA}
    \label{fig:A4_7}
  \end{subfigure}
  \hfill
  \begin{subfigure}{0.24\linewidth}
    \includegraphics[width=\textwidth]{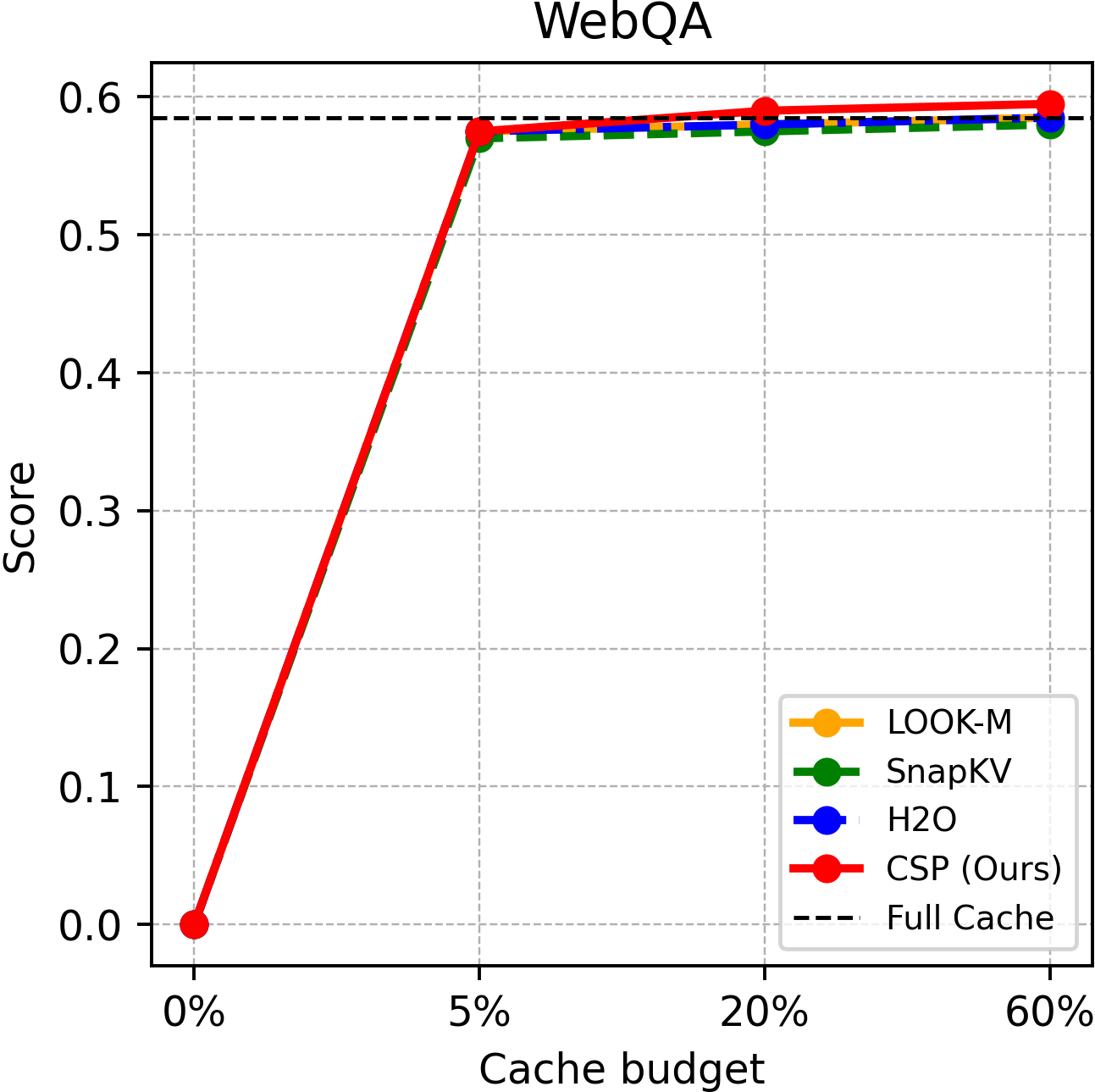}
    \caption{WebQA}
    \label{fig:A4_8}
  \end{subfigure}
  \caption{The impact of the cache size budget.}
  \label{fig:A4}
  \vspace{-3mm}
\end{figure*}

\subsection{Influence of Model Architectures}
\label{sub:sec}
In this section, we evaluate the influence of model architecture on the performance of our proposed KV cache method across selected tasks (T-2, T-4, S-4, and IR) in the benchmark. We introduce two more architectures: InternVL-v1.5-7B \cite{chen2024internvl}, which scales up the vision encoder with cross-modal integration, and MobileVLM-V2-3B \cite{chu2024mobilevlm}, which features a lighter structure with an efficient downsampling projector (LDPv2) and focuses on intra-modal processing. Table \ref{tab:A5} demonstrates our test results of our methods compared with baselines. For InternVL-v1.5-7B, we observe that CSP achieves the highest performance in most tasks, particularly in T-2 (22.8) and S-4 (25.4), indicating that our KV cache pruning method benefits from InternVL’s large-scale vision encoder. This architecture supports a robust cross-modal alignment, which CSP leverages by retaining crucial tokens independently from cross-attention and self-attention regions, maintaining contextual richness in visual-textual integration tasks. Regarding MobileVLM-V2-3B, CSP also demonstrates superior performance, especially in IR (5.3), where precise image retrieval benefits from MobileVLM’s lightweight, modality-aware processing enabled by LDPv2. The efficiency of this architecture allows CSP to perform well by focusing on high-salience tokens in the vision domain, as seen in improved scores across tasks. This shows that MobileVLM’s intra-modal alignment complements CSP’s motivations of avoiding the collapse of token selection.

\begin{table}[h!]
\centering
\begin{tabular}{lcccc}
\toprule
\textbf{Method} & \textbf{T-2} & \textbf{T-4} & \textbf{S-4} & \textbf{IR} \\
\midrule
\hdashline 
\multicolumn{5}{c}{\textit{InternVL-v1.5-7B} \cite{chen2024internvl}} \\
\hdashline 
Full Cache & 19.2 & 21.3 & 19.1 & 0.0 \\
H$_2$O \cite{zhang2024h2o} & 20.0 & 20.4 & 19.6 & 0.5 \\
SnapKV \cite{li2024snapkv} & 19.9 & 19.5 & 19.4 & 0.2 \\
RoCo \cite{ren2024efficacy} & 20.0 & 18.5 & 19.6 & 0.5 \\
LOOK-M \cite{ren2024efficacy} & 22.0 & 19.6 & 22.9 & 0.5 \\ 
CSP (Ours) & \textbf{22.8} & \textbf{20.7} & \textbf{25.4} & \textbf{0.6} \\ 
\hdashline 
\multicolumn{5}{c}{\textit{MobileVLM-V2-3B}\cite{chu2024mobilevlm}} \\
\hdashline 
Full Cache & 46.2 & 38.5 & 33.0 & 4.7 \\
H$_2$O \cite{zhang2024h2o} & 46.4 & 38.2 & 28.2 & 4.5 \\
SnapKV \cite{li2024snapkv} & 46.4 & 38.5 & 27.2 & 4.7 \\
RoCo \cite{ren2024efficacy} & 46.6 & 38.0 & 28.9 & 4.6 \\
LOOK-M \cite{wan2024look} & 47.0 & 38.7 & 32.8 & 4.8 \\
CSP (Ours) & \textbf{47.3} & \textbf{39.0} & \textbf{33.1} & \textbf{5.3} \\ 
\bottomrule
\end{tabular}
\caption{Comparison of KV cache methods across tasks (T-2, T-4, S-4, IR) on InternVL-v1.5-7B \cite{chen2024internvl} and MobileVLM-V2-3B \cite{chu2024mobilevlm}.
}
\vspace{-6mm}
\label{tab:A5}
\end{table}


\section{Conclusion}
\label{sec:conclusion}
In this work, we propose Cross-Self Pruning (CSP), a simple and training-free KV cache method designed to independently select top-k tokens from cross-attention and self-attention regions. 
We evaluate pruning method on a range of multimodal tasks, demonstrating that CSP achieves competitive performance across all tasks while reducing cache budgets. Furthermore, our ablation study highlights the method's robustness and effectiveness in optimizing token selection through intra- and cross-modality pruning, offering a lightweight solution for improving computational efficiency without compromising model accuracy.

{\small
\bibliographystyle{ieeenat_fullname}
\bibliography{11_references}
}

\ifarxiv \clearpage \appendix \section{Appendix Section}
\label{sec:appendix_section}
\subsection{Ratio Selection for Dataset}
The table reveals varying strategies for attention configuration across datasets, reflecting task-specific priorities. Datasets like \textit{Spot-the-Diff} and \textit{WebQA} emphasize cross-attention by assigning 90\% of the top-k selection to cross-modal interactions. Conversely, tasks such as \textit{ActionPrediction} rely entirely on intra-modal attention, with no top-k selection allocated to cross-attention. Overall, most datasets adopt a balanced approach, allocating 50\% of the top-k selection to cross-attention and maintaining a recency bias of 1, indicating a general preference for equal weighting of intra- and cross-modal attention in multi-modal inference.

\begin{table}[ht]
\centering
\begin{tabular}{lcc}
\toprule
\textbf{Dataset}          & \textit{LLaVA-7B} & \textit{LLaVA-13B} \\ \midrule
T1                   & 0.5                            & 1                                  \\ 
T2                   & 0.5                            & 0.5                                \\ 
T3                   & 0.5                            & 0.5                                \\ 
T4                   & 0.1                            & 0.5                                \\ 
S1                   & 0.5                            & 0.5                                \\ 
S2                   & 0.5                            & 0.5                                \\ 
S3                   & 0.5                            & 0.5                                \\ 
S4                   & 0.5                            & 0.5                                \\ 
NH                   & 0.5                            & 0.5                                \\ 
IR                   & 0.9                            & 0.9                                \\ \bottomrule
\end{tabular}
\caption{Cross Ratio Selection for Different Tasks.}
\label{tab:dataset_ratios}
\vspace{-3mm}
\end{table}

\subsection{Distribution Visualization}
In this section, we visualize the distribution differences between intra- and cross-attention using kernel density estimation (KDE). From the visualizations, we observe that certain datasets exhibit a stronger reliance on cross-attention, while others depend more heavily on self-attention.
Here we apply the Kernel Density Estimation (KDE) and Jensen-Shannon (JS) divergence to analysis the difference of distributions.

In terms of KDE, figure \ref{fig:app_kde} demonstrates the attention weight distributions for both self-attention (blue) and cross-attention (red) across eight datasets. It is evident that the two types of attention exhibit distinct patterns depending on the dataset, which directly impacts the pruning strategies during the KV cache process. For datasets such as \textit{CLEVR-Change} and \textit{CounterfactualInference}, cross-attention weights show a significantly concentrated and dominant peak at very low values, while self-attention demonstrates broader coverage. This suggests that cross-attention contributes heavily to the model's decision-making in these tasks, emphasizing token dependencies between modalities (e.g., image-text). Pruning strategies in these cases might inadvertently eliminate crucial cross-attention connections, leading to incomplete information transfer and subsequent degradation in inference accuracy. Conversely, datasets such as \textit{DocVQA} and \textit{EgocentricNavigation} reveal more dispersed and substantial self-attention weights, while cross-attention peaks remain narrow. This indicates a reliance on intra-modal token interactions, such as contextual reasoning within the same modality. Aggressive pruning of self-attention tokens in such cases risks losing key intra-modal context, adversely impacting downstream predictions. 

To further quantify the discrepancy between self-attention and cross-attention distributions, we compute the Jensen-Shannon (JS) divergence for each dataset. Higher divergence values suggest a stark imbalance between the two attention mechanisms, indicating that naive pruning may disproportionately affect one type of attention. In Figure \ref{fig:app_js}, tasks like \textit{CLEVR-Change} and \textit{ActionPrediction} exhibit high divergence, implying the necessity for task-specific pruning thresholds to retain balanced contributions from both self- and cross-attention, whereas tasks like \textit{DocVQA} show lower divergence, where self- and cross-attention operate more harmoniously, and uniform pruning strategies may suffice. These distribution discrepancies highlight several challenges and insights. Uniform pruning approaches that prioritize magnitude-based filtering may disproportionately affect regions with dense cross-attention peaks (e.g., \textit{CLEVR-Change}), hindering the model's ability to encode cross-modal dependencies, particularly in visual reasoning tasks. For datasets where self-attention dominates (e.g., \textit{EgocentricNavigation}), pruning strategies that overly prioritize low-weight tokens may reduce contextual coherence, especially for long-context sequences. The analysis underscores the necessity for adaptive pruning mechanisms that balance retention across self- and cross-attention regions, guided by the specific reliance patterns observed in the KDE and JS divergence results. Furthermore, the differential pruning of self- and cross-attention tokens influences the effectiveness of KV cache utilization. Over-pruning either region may cause a skewed representation in the cache, reducing its utility for long-context inference. 


The observed disparities in self- and cross-attention distributions across datasets necessitate an adaptive pruning framework that dynamically adjusts based on task-specific ratio, effectively balancing the preservation of essential intra- and cross-modal dependencies while optimizing KV cache utilization.

\begin{figure*}[!htb]
  \centering
  \begin{subfigure}{0.24\linewidth}
    \includegraphics[width=\textwidth]{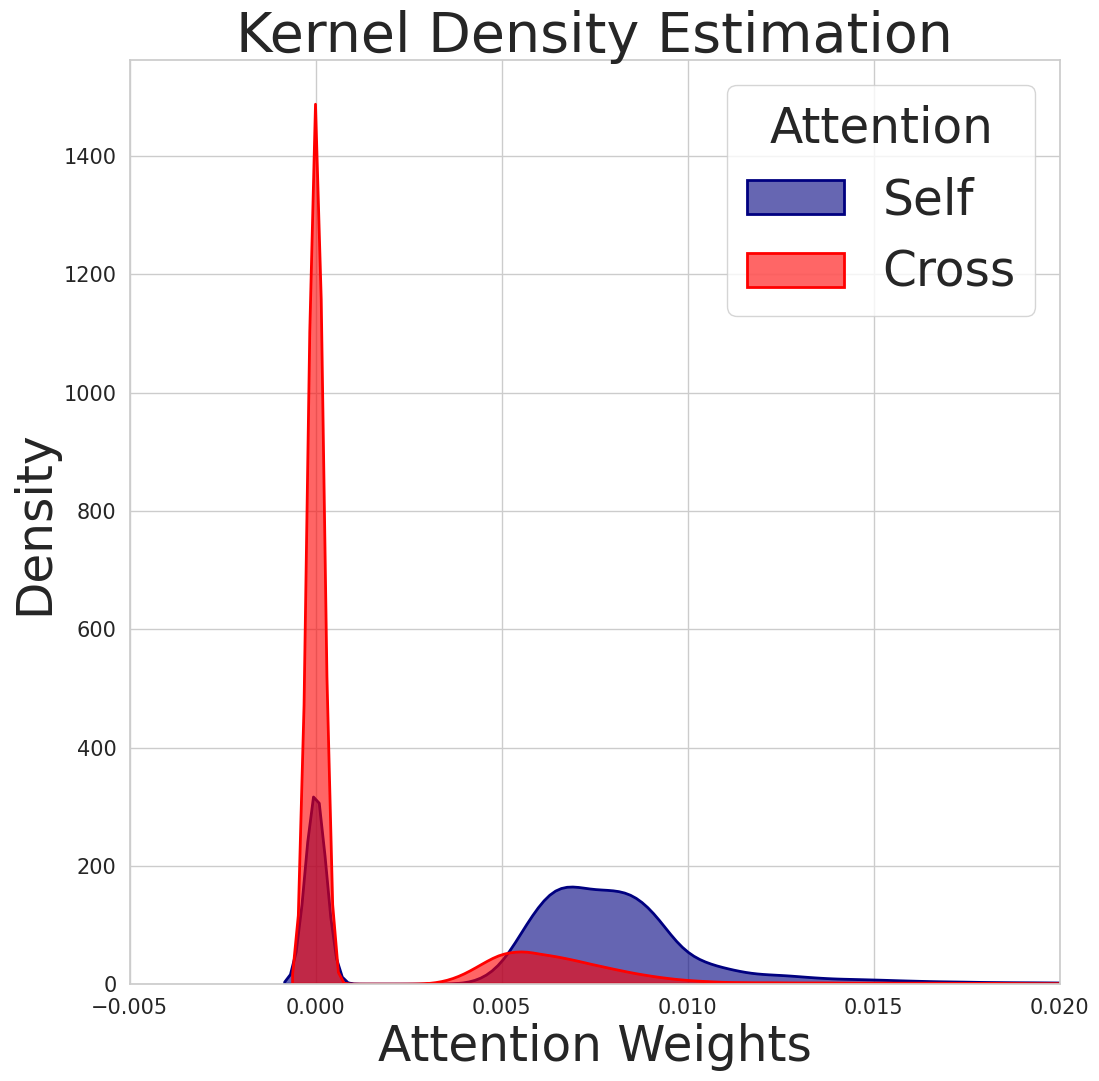} 
    \caption{ActionLocalization}
    \label{fig:A_ActionLocalization}
  \end{subfigure}
  \hfill
  \begin{subfigure}{0.24\linewidth}
    \includegraphics[width=\textwidth]{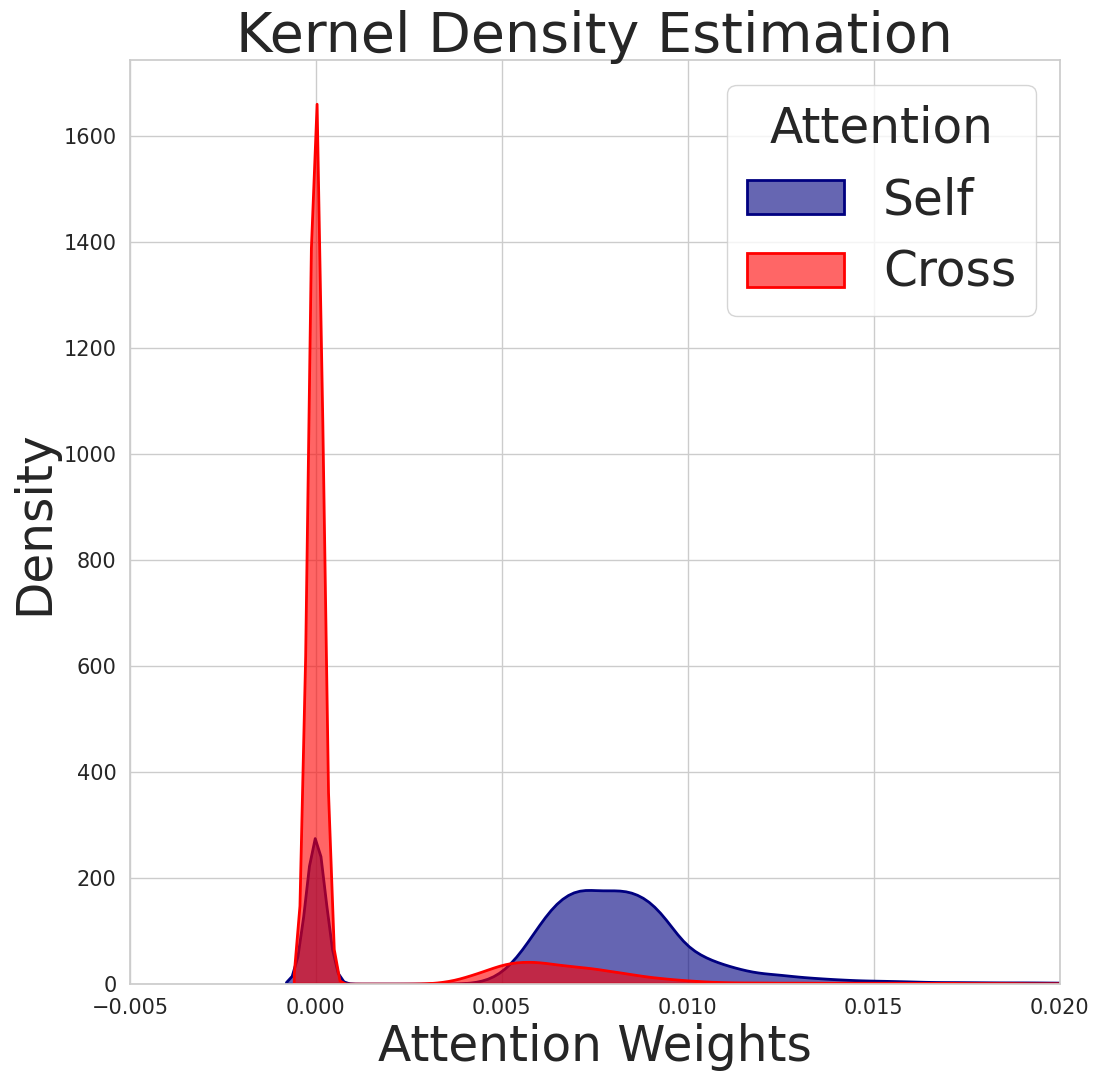}
    \caption{ActionPrediction}
    \label{fig:A_ActionPrediction}
  \end{subfigure}
  \hfill
  \begin{subfigure}{0.24\linewidth}
    \includegraphics[width=\textwidth]{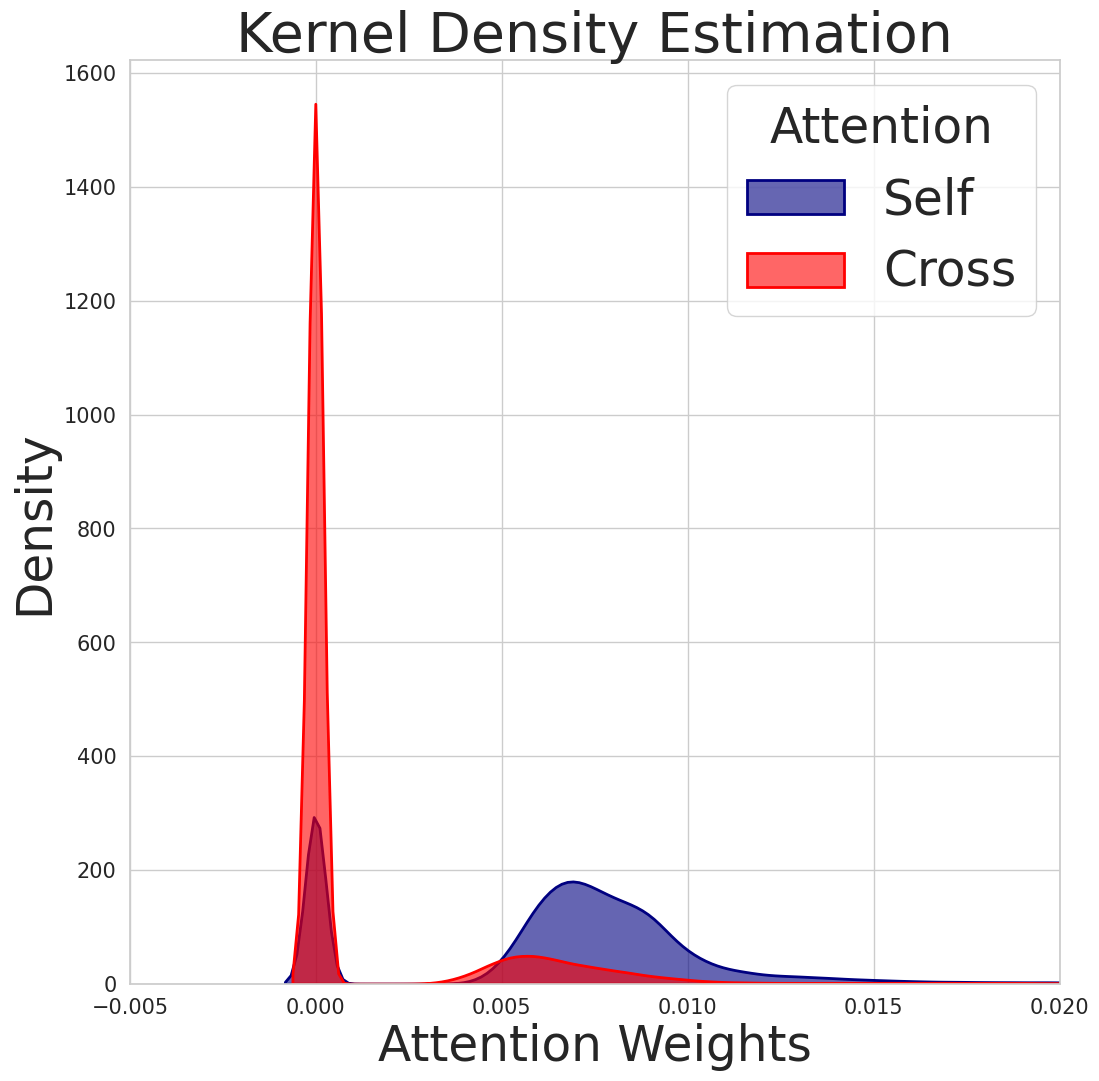}
    \caption{ActionSequence}
    \label{fig:A_ActionSequence}
  \end{subfigure}
  \hfill
  \begin{subfigure}{0.24\linewidth}
    \includegraphics[width=\textwidth]{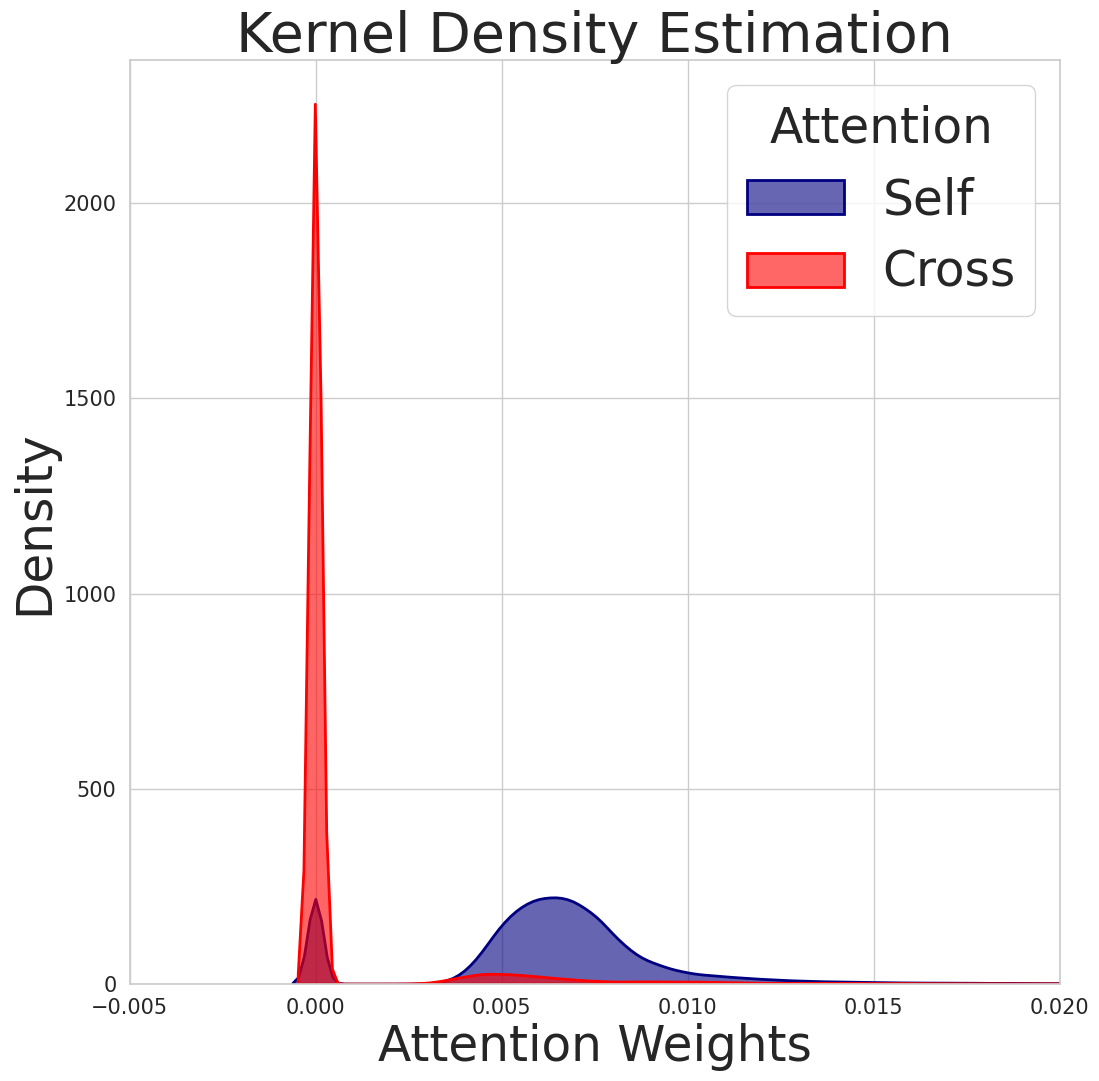} 
    \caption{CharacterOrder}
    \label{fig:A_CharacterOrder}
  \end{subfigure}
  \hfill
  \begin{subfigure}{0.24\linewidth}
    \includegraphics[width=\textwidth]{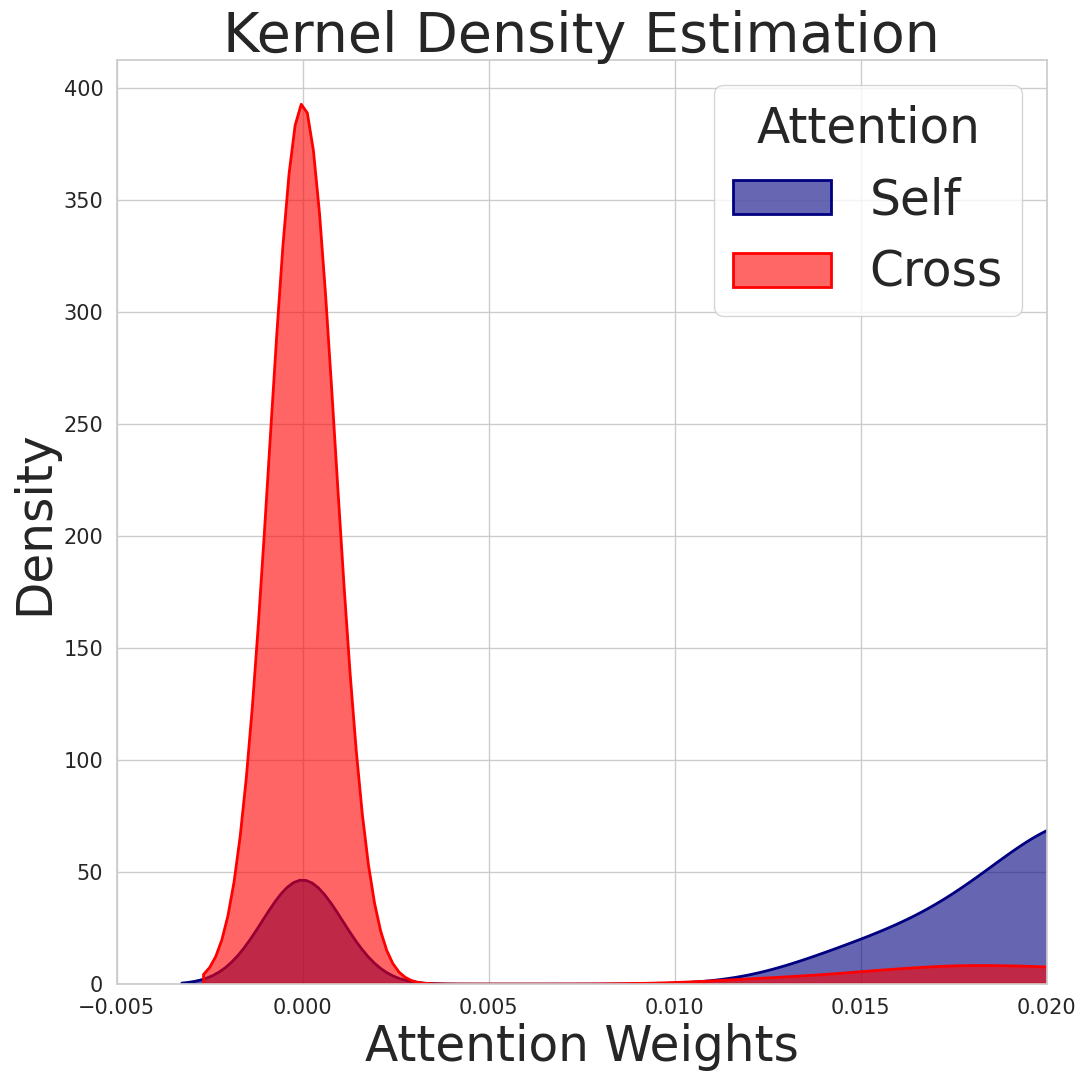}
    \caption{CLEVR-Change}
    \label{fig:A_CLEVR-Change}
  \end{subfigure}
  \hfill
  \begin{subfigure}{0.24\linewidth}
    \includegraphics[width=\textwidth]{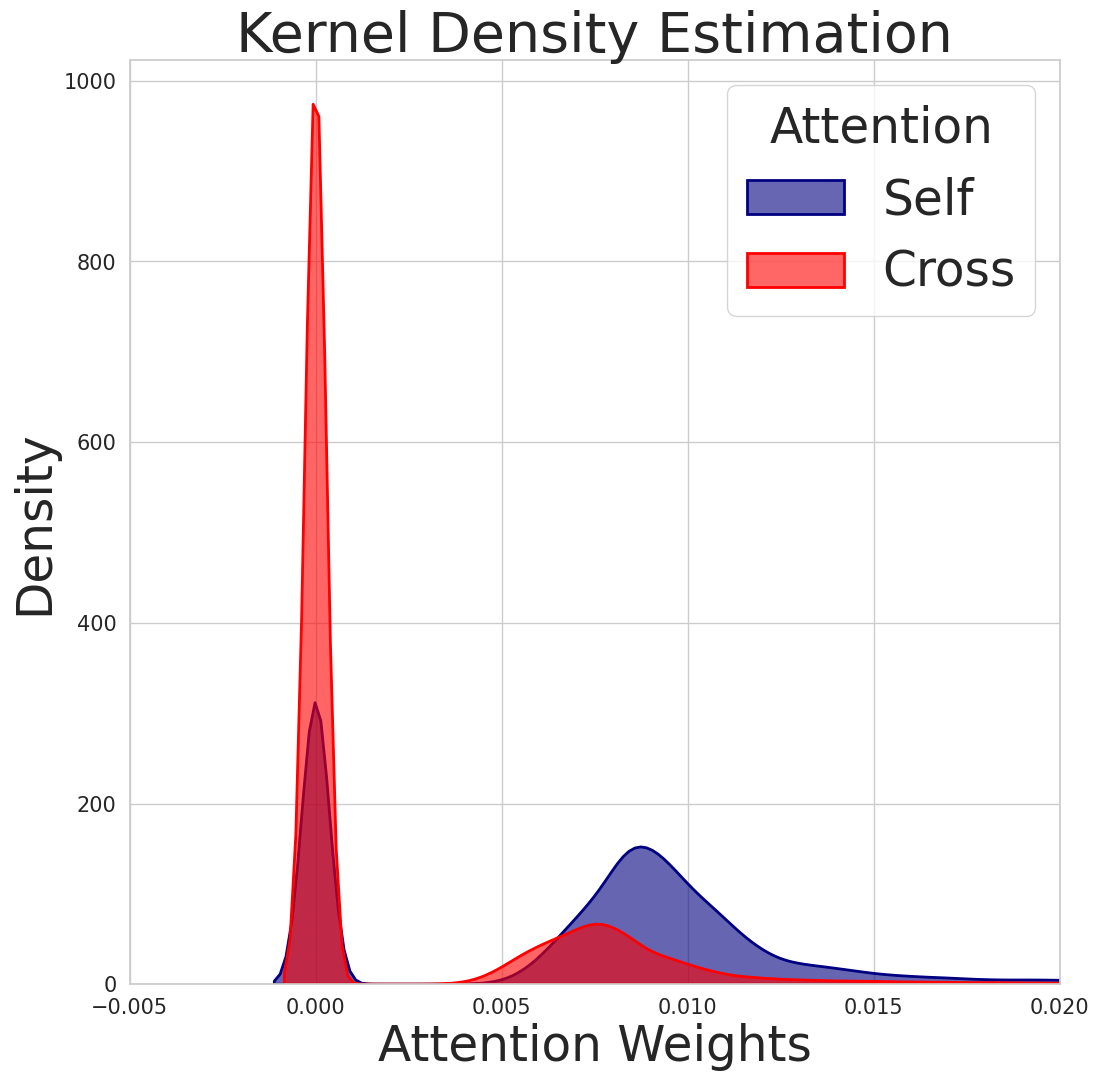}
    \caption{CounterfactualInference}
    \label{fig:A_CounterfactualInference}
  \end{subfigure}
  \hfill
  \begin{subfigure}{0.24\linewidth}
    \includegraphics[width=\textwidth]{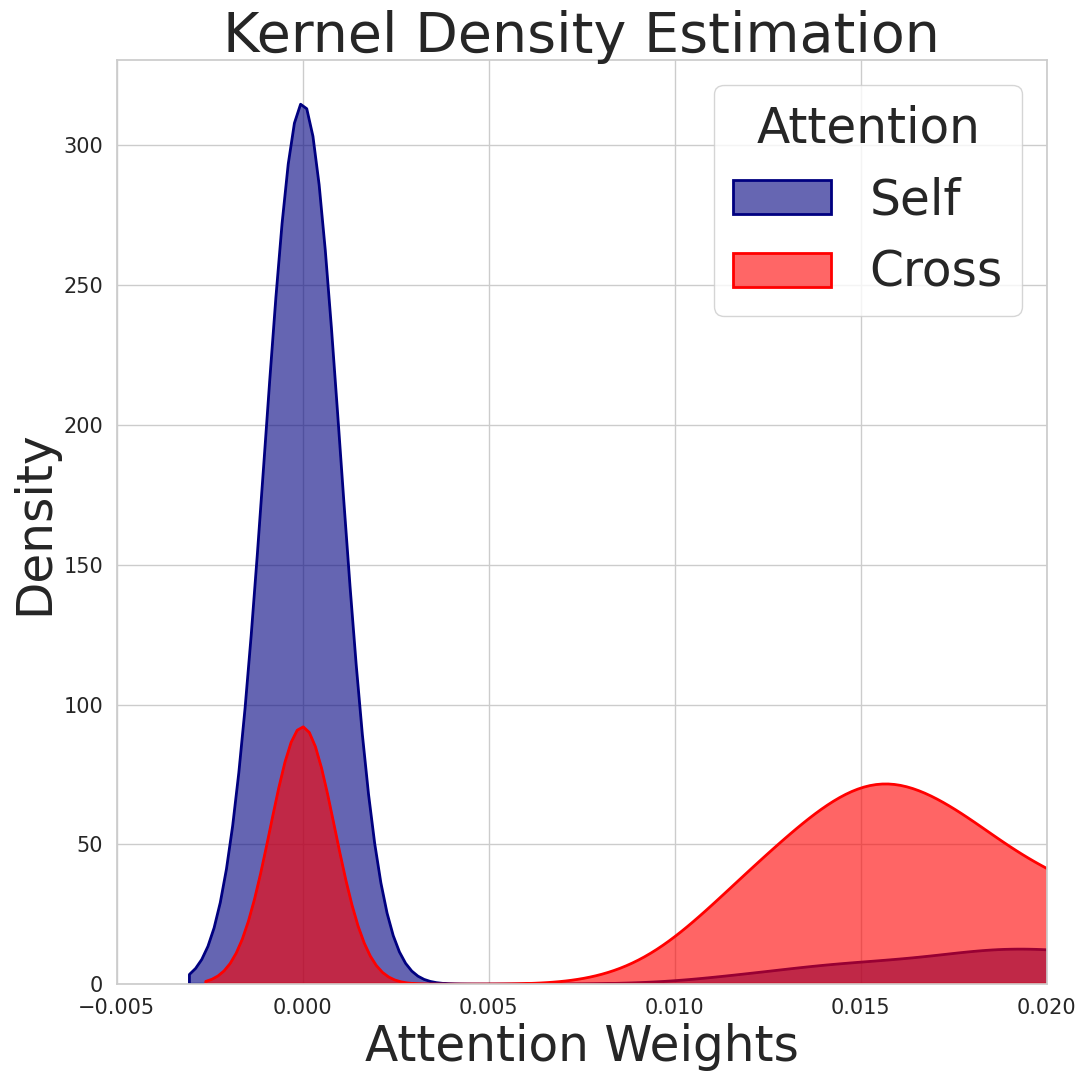}
    \caption{DocVQA}
    \label{fig:A_DocVQA}
  \end{subfigure}
  \hfill
  \begin{subfigure}{0.24\linewidth}
    \includegraphics[width=\textwidth]{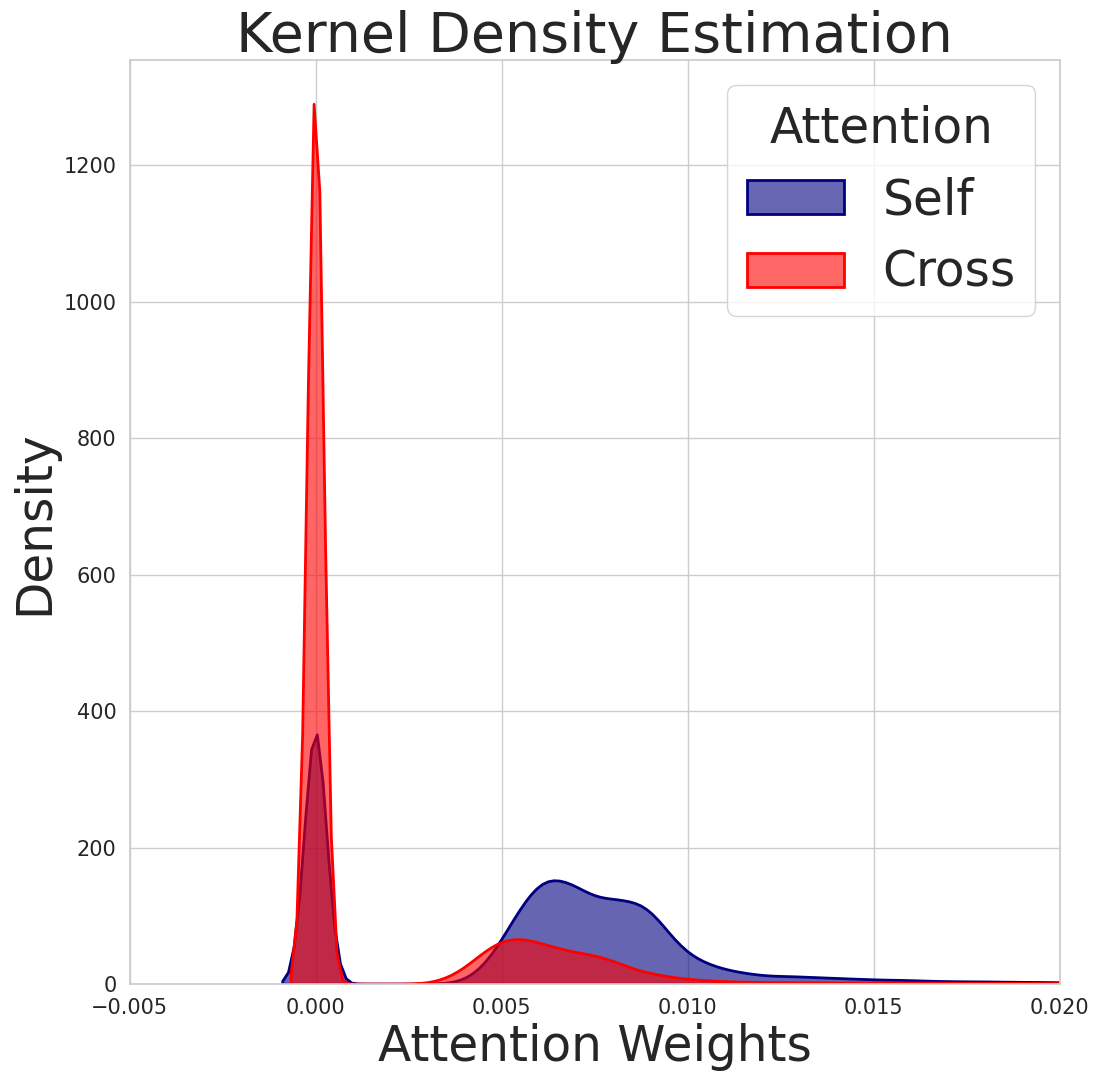}
    \caption{EgocentricNavigation}
    \label{fig:A_EgocentricNavigation}
  \end{subfigure}
  \caption{Kernel Density Estimation (KDE) of the attention weight distributions.}
  \label{fig:app_kde}
  \vspace{-3mm}
\end{figure*}

\begin{figure*}[!th]
  \centering
  \begin{subfigure}{0.24\linewidth}
    \includegraphics[width=\textwidth]{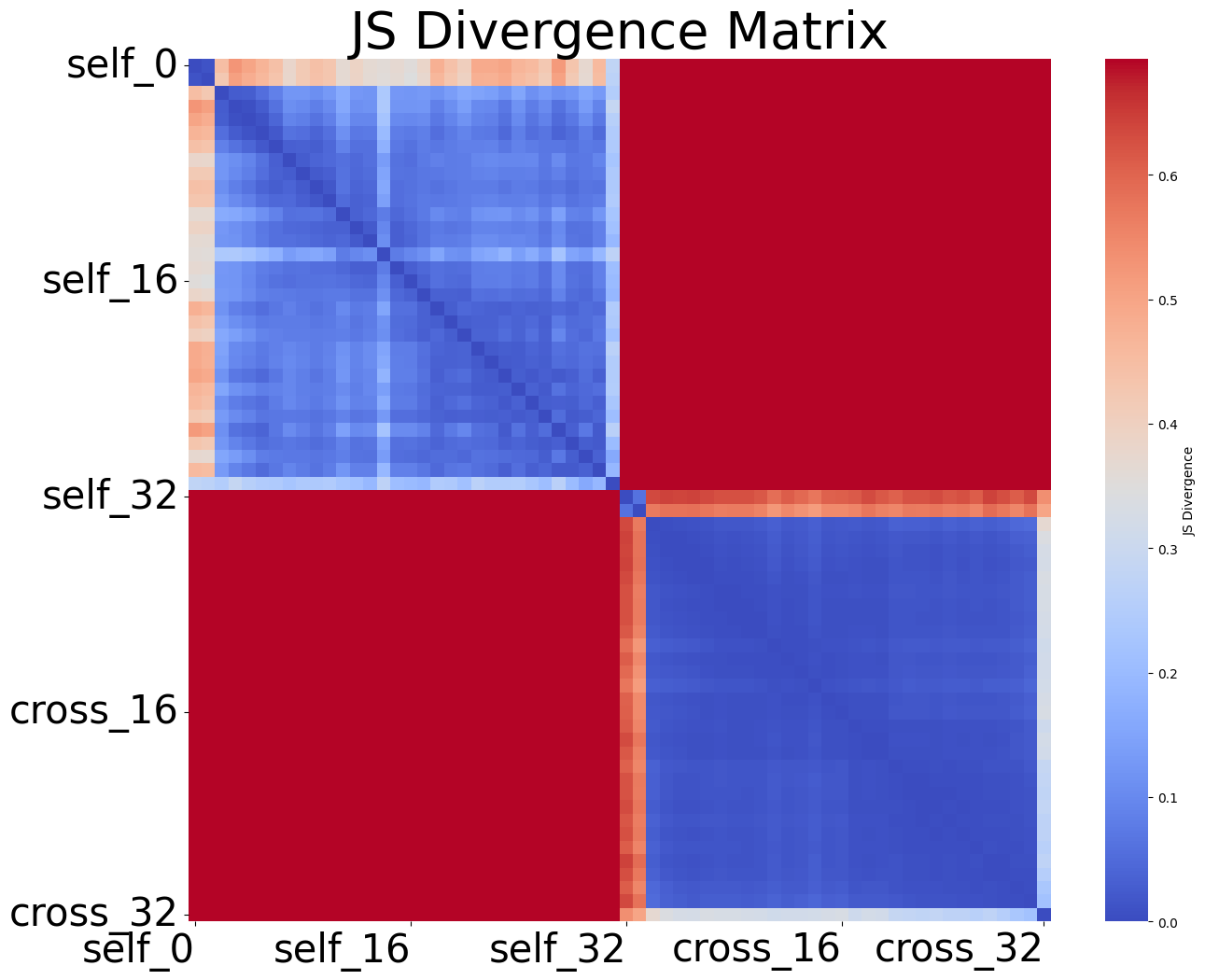} 
    \caption{ActionLocalization}
    \label{fig:AP_1}
  \end{subfigure}
  \hfill
  \begin{subfigure}{0.24\linewidth}
    \includegraphics[width=\textwidth]{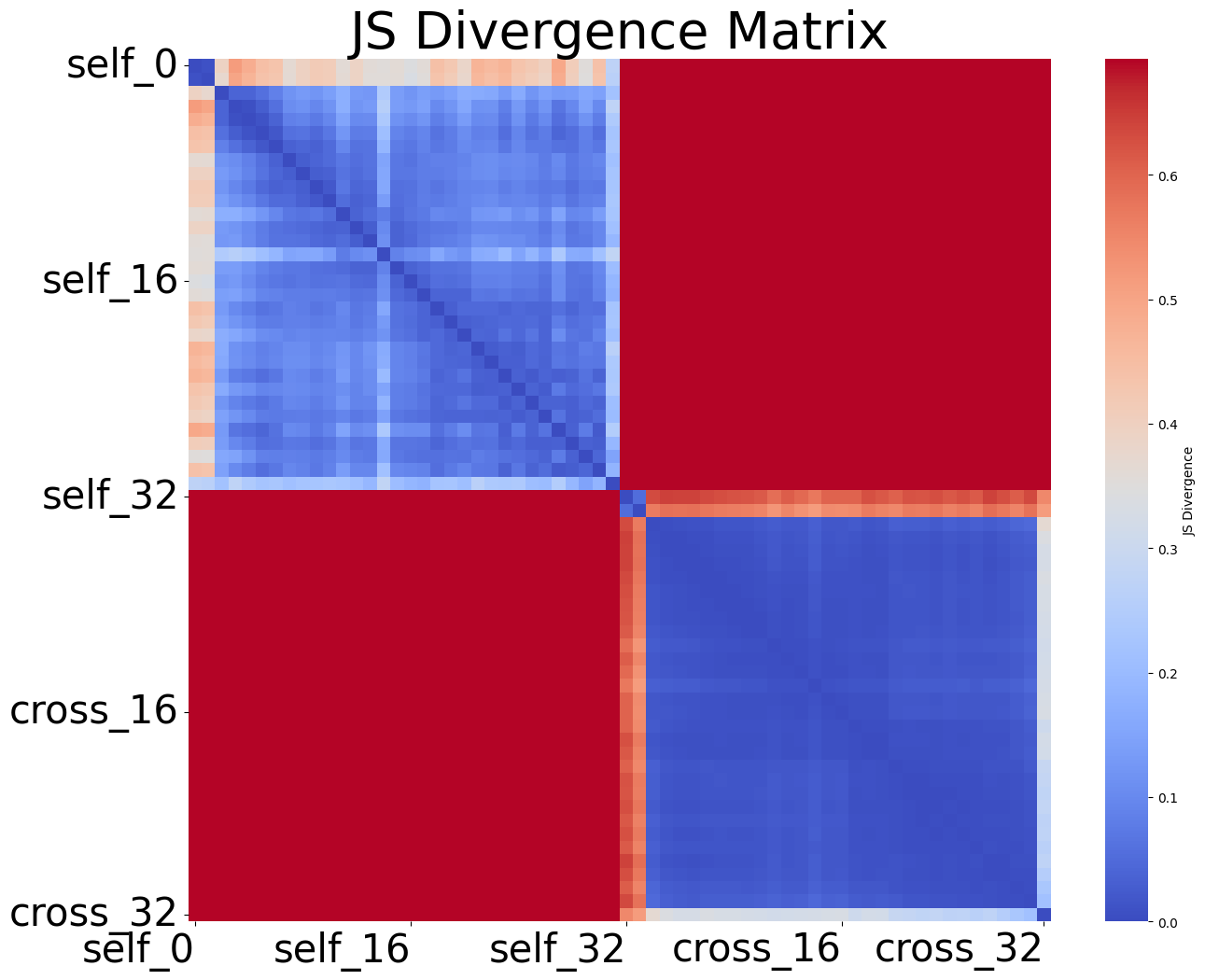}
    \caption{ActionPrediction}
    \label{fig:AP_2}
  \end{subfigure}
  \hfill
  \begin{subfigure}{0.24\linewidth}
    \includegraphics[width=\textwidth]{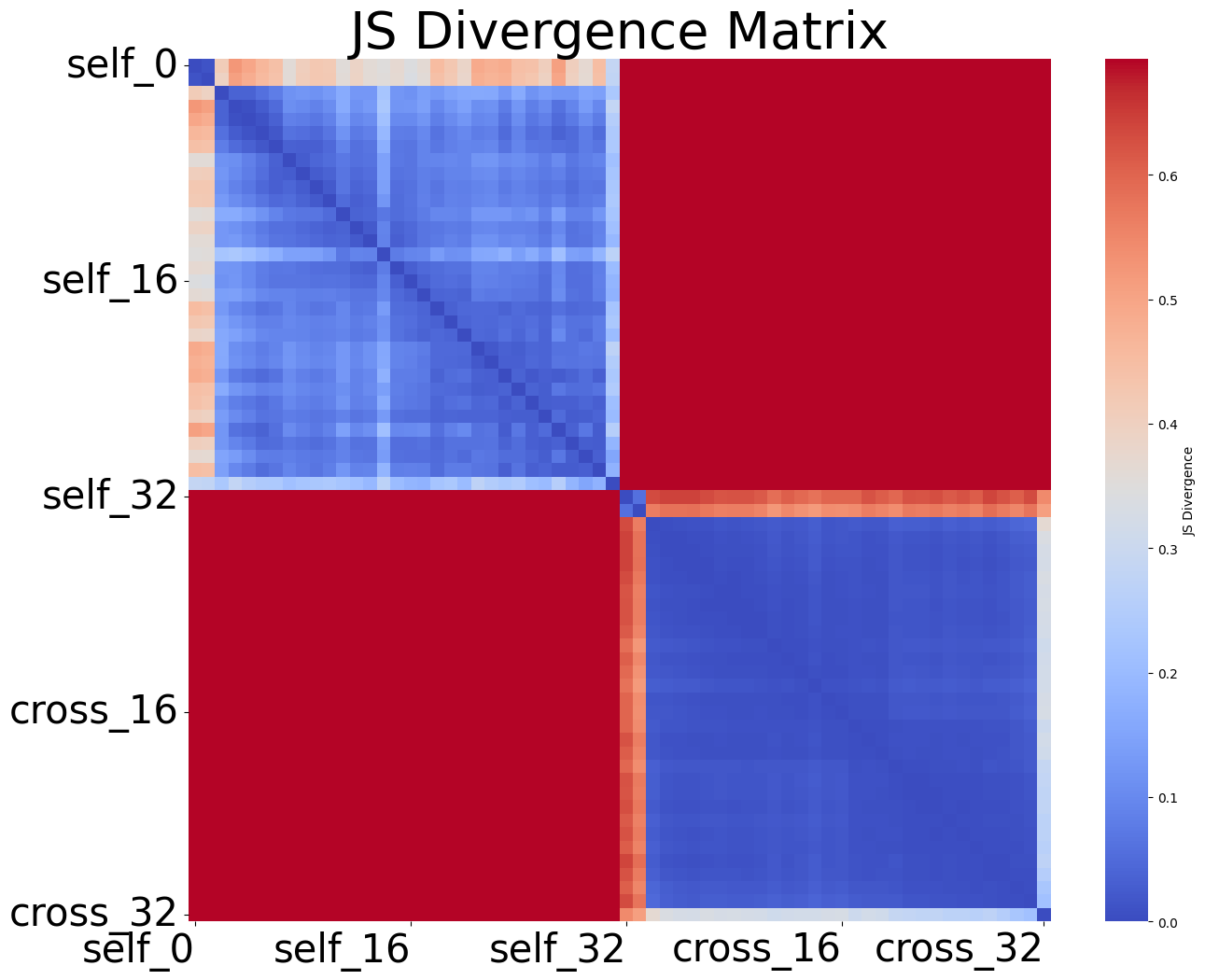}
    \caption{ActionSequence}
    \label{fig:AP_3}
  \end{subfigure}
  \hfill
  \begin{subfigure}{0.24\linewidth}
    \includegraphics[width=\textwidth]{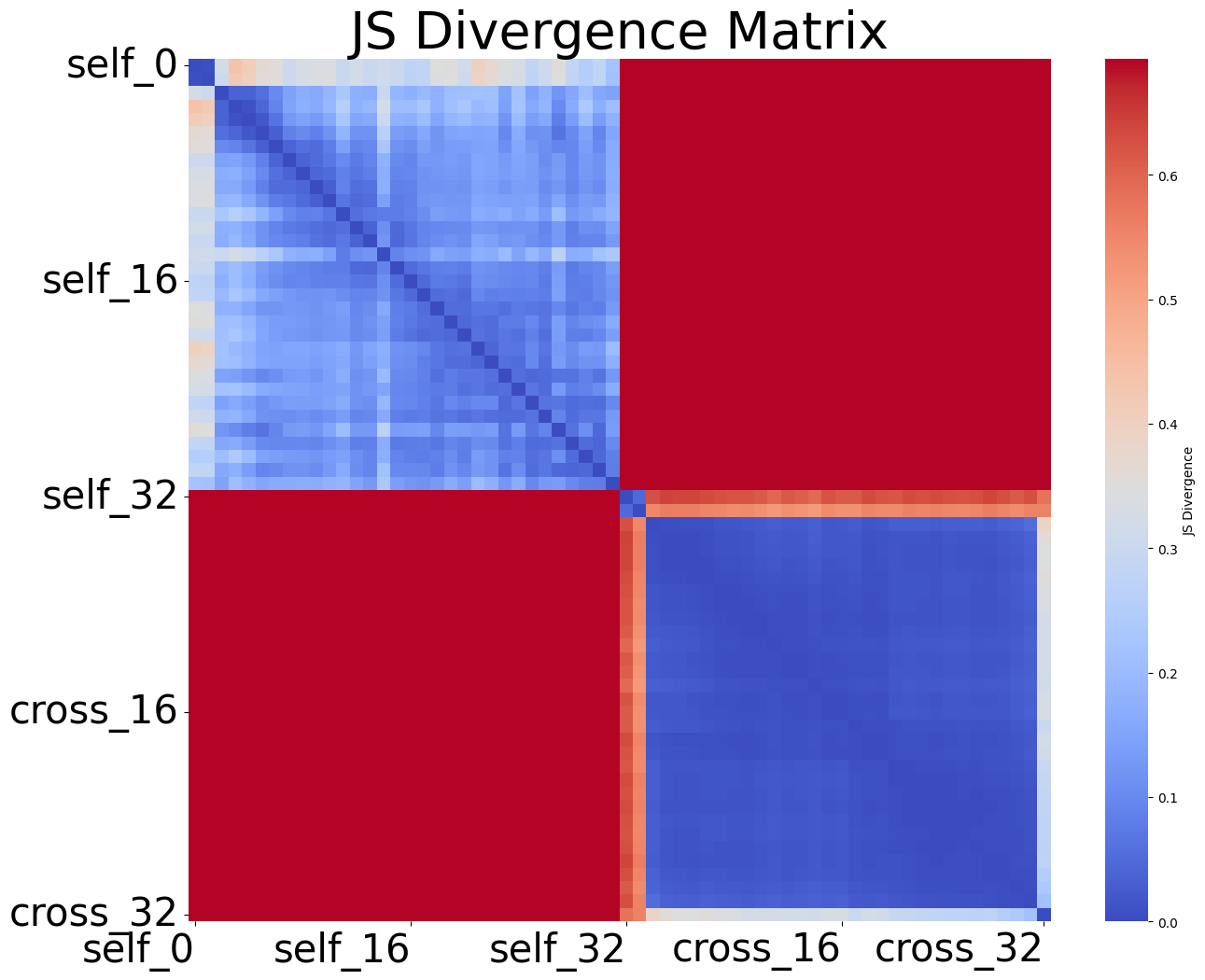}
    \caption{ALFRED}
    \label{fig:AP_4}
  \end{subfigure}
  \begin{subfigure}{0.24\linewidth}
    \includegraphics[width=\textwidth]{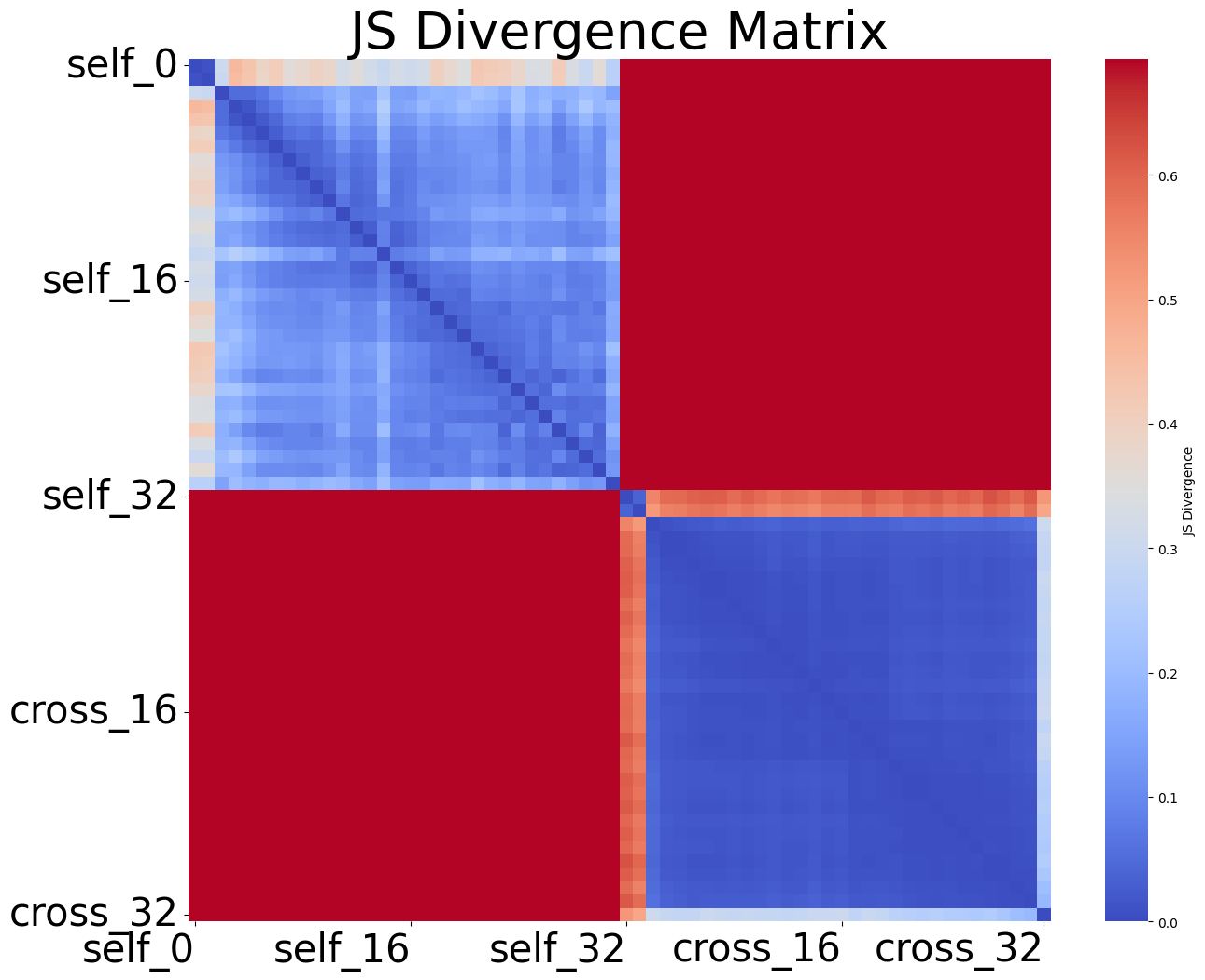} 
    \caption{CharacterOrder}
    \label{fig:AP_5}
  \end{subfigure}
  \hfill
  \begin{subfigure}{0.24\linewidth}
    \includegraphics[width=\textwidth]{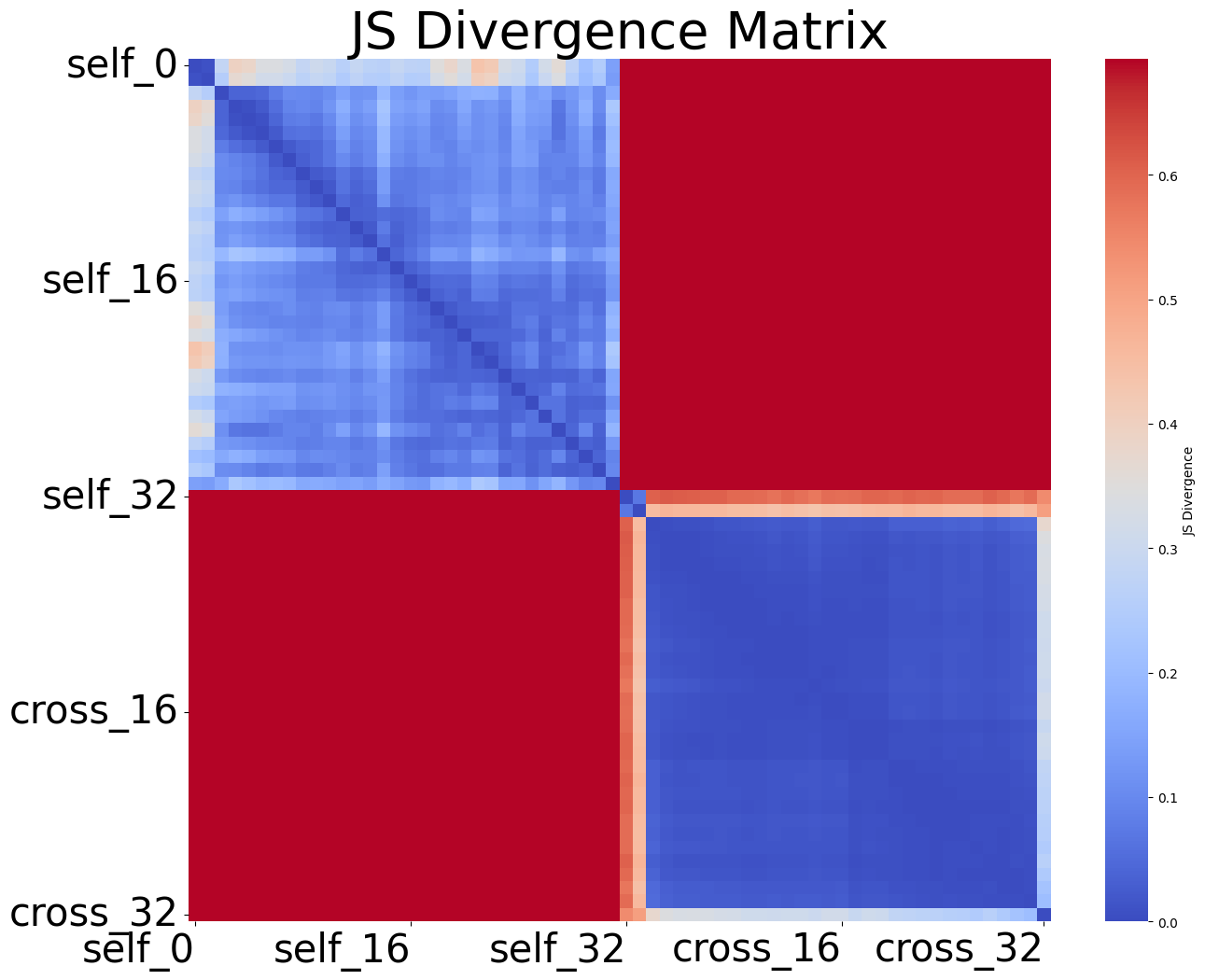}
    \caption{CLEVR-Change}
    \label{fig:AP_6}
  \end{subfigure}
  \hfill
  \begin{subfigure}{0.24\linewidth}
    \includegraphics[width=\textwidth]{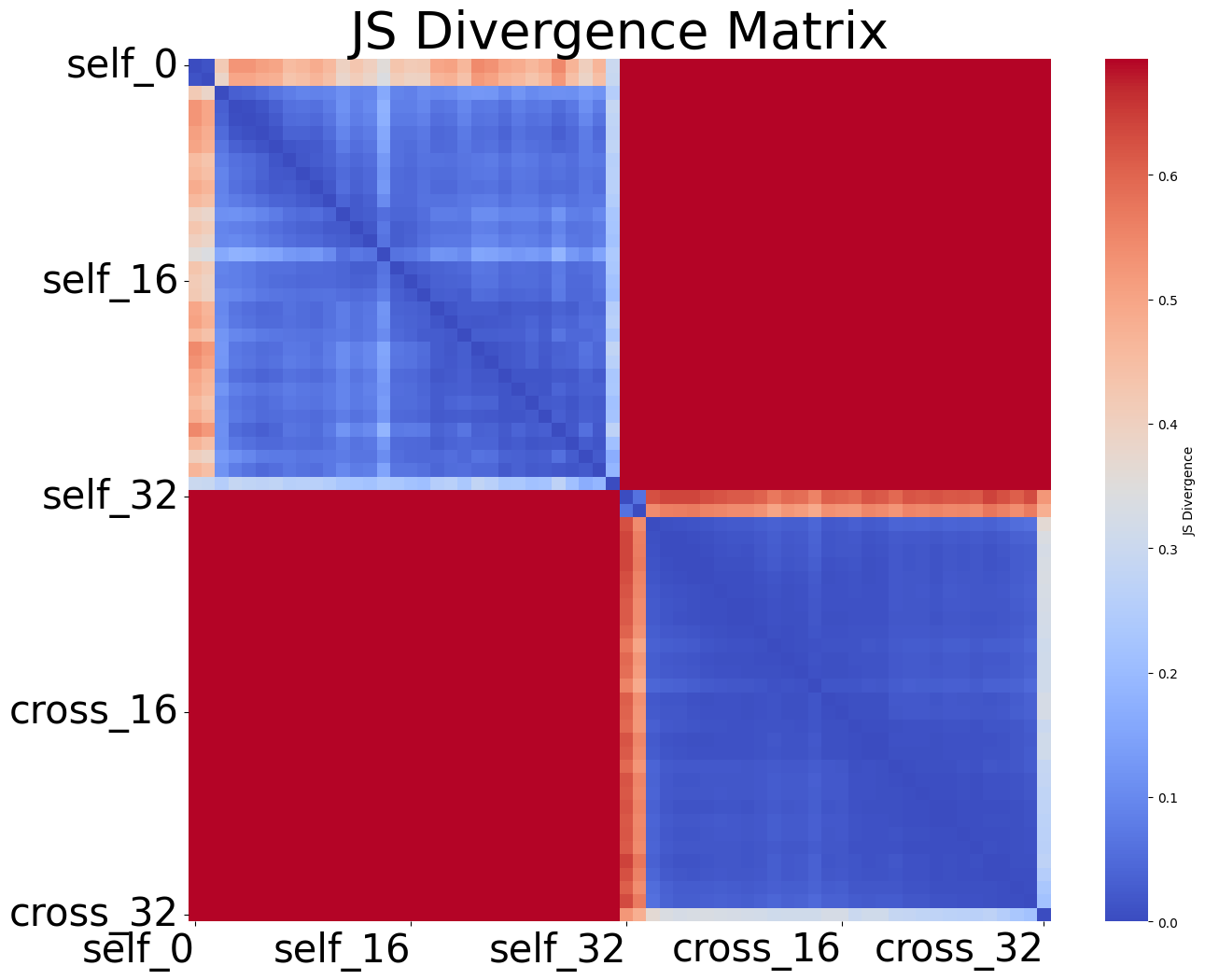}
    \caption{CounterfactualInference}
    \label{fig:AP_7}
  \end{subfigure}
  \hfill
  \begin{subfigure}{0.24\linewidth}
    \includegraphics[width=\textwidth]{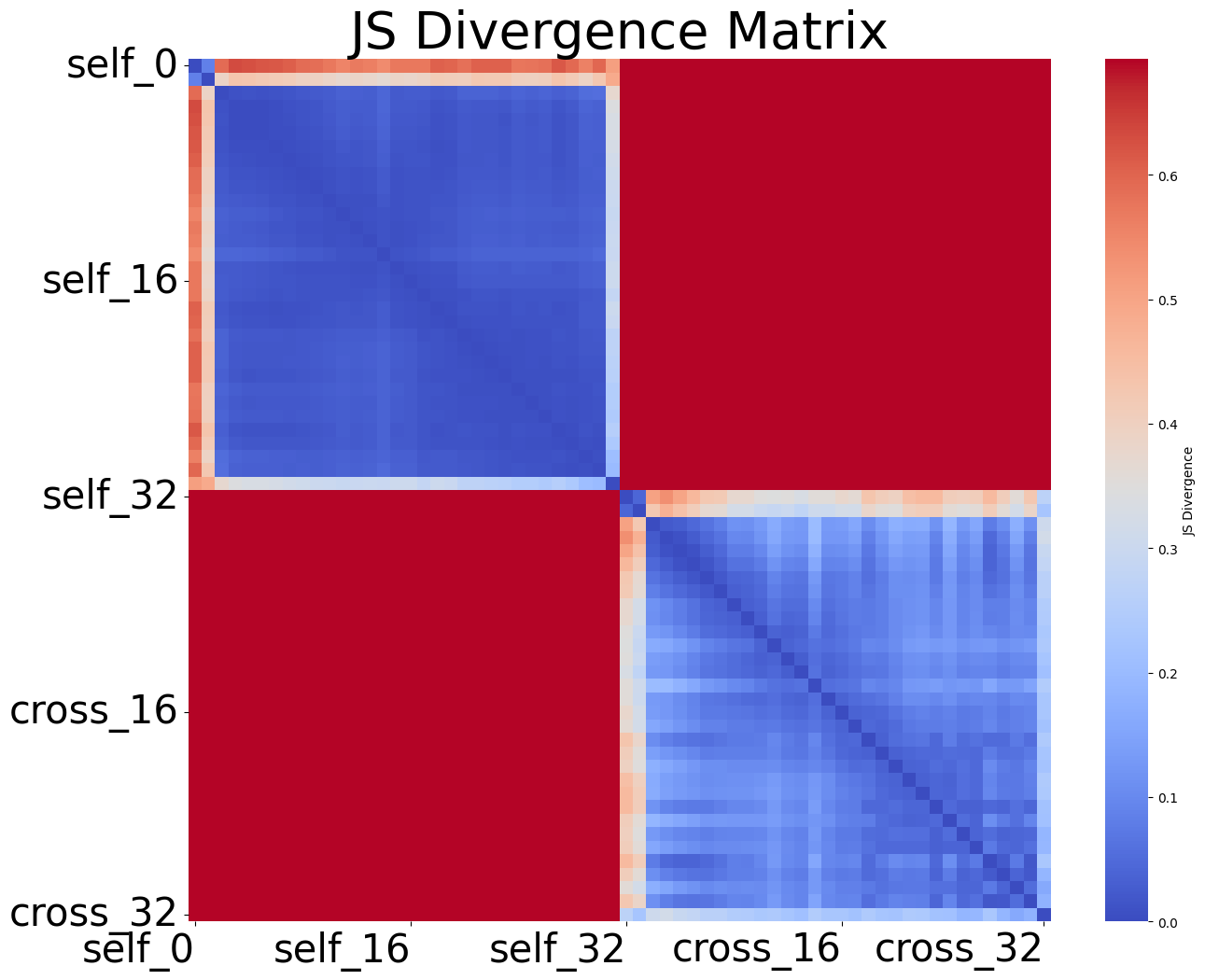}
    \caption{DocVQA}
    \label{fig:AP_8}
  \end{subfigure}
  \caption{Jensen-Shannon (JS) divergence scores between cross-attention and self-attention across all layers.}
  \label{fig:app_js}
  \vspace{-3mm}
\end{figure*}

\begin{table*}[htb]
\centering
\caption{Detailed Statistics and Taxonomy from \textbf{MILEBench} \cite{song2024milebench}.}
\label{tab:milebench}
\resizebox{\textwidth}{!}{%
\begin{tabular}{|c|c|c|c|c|c|}
\hline
\textbf{Category} & \textbf{Task} & \textbf{Dataset} & \textbf{Data Source} & \textbf{Count} & \textbf{Metric} \\ \hline
\multirow{4}{*}{Temporal Multi-image} & Action Understanding and Prediction (T-1) & \begin{tabular}[c]{@{}c@{}}Action Localization \\ Action Prediction \\ Action Sequence\end{tabular} & \begin{tabular}[c]{@{}c@{}}STA \cite{gao2017tall} \\ STAR \cite{wu2024star} \\ STAR \cite{wu2024star}\end{tabular} & 200 & Accuracy \\ \cline{2-6} 
 & Object and Scene Understanding (T-2) & \begin{tabular}[c]{@{}c@{}}Object Existence \\ Object Interaction \\ Moving Attribute \\ Object Shuffle\end{tabular} & \begin{tabular}[c]{@{}c@{}}CLEVRER \cite{yi2019clevrer} \\ STAR \cite{wu2024star} \\ CLEVRER \cite{yi2019clevrer} \\ Perception Test \cite{patraucean2024perception}\end{tabular} & 200 & Accuracy \\ \cline{2-6} 
 & Visual Navigation and Spatial Localization (T-3) & \begin{tabular}[c]{@{}c@{}}Egocentric Navigation \\ Moving Direction\end{tabular} & \begin{tabular}[c]{@{}c@{}}VLN-CE \cite{krantz2020beyond} \\ CLEVRER \cite{yi2019clevrer}\end{tabular} & 200 & Accuracy \\ \cline{2-6} 
 & Counterfactual Reasoning and State Change (T-4) & \begin{tabular}[c]{@{}c@{}}Counterfactual Inference \\ State Change \\ Character Order \\ Scene Transition\end{tabular} & \begin{tabular}[c]{@{}c@{}}CLEVRER \cite{yi2019clevrer} \\ Perception Test \cite{patraucean2024perception} \\ Perception Test \cite{patraucean2024perception} \\ MovieNet \cite{huang2020movienet}\end{tabular} & 200 & Accuracy \\ \hline
\multirow{5}{*}{Semantic Multi-image} & Knowledge Grounded QA (S-1) & \begin{tabular}[c]{@{}c@{}}Webpage QA \\ Textbook QA \\ Complex Multimodal QA \\ Long Text with Images QA\end{tabular} & \begin{tabular}[c]{@{}c@{}}WebQA \cite{chang2022webqa} \\ TQA \cite{kembhavi2017you} \\ MultiModalQA \cite{talmor2021multimodalqa} \\ WikiVQA\end{tabular} & 200 & Accuracy \\ \cline{2-6} 
 & Text-Rich Images QA (S-2) & \begin{tabular}[c]{@{}c@{}}Slide QA \\ OCR QA \\ Document QA\end{tabular} & \begin{tabular}[c]{@{}c@{}}SlideVQA \cite{tanaka2023slidevqa} \\ OCR-VQA \cite{mishra2019ocr} \\ DocVQA \cite{mathew2021docvqa}\end{tabular} & 200 & Accuracy \\ \cline{2-6} 
 & Visual Relation Inference (S-3) & \begin{tabular}[c]{@{}c@{}}Visual Change Captioning \\ Visual Relationship Expressing\end{tabular} & \begin{tabular}[c]{@{}c@{}}Spot-the-Diff \cite{jhamtani2018learning} \\ CLEVR-Change \cite{hosseinzadeh2021image}\end{tabular} & 200 & ROUGE-L \\ \cline{2-6} 
 & Dialogue (S-4) & \begin{tabular}[c]{@{}c@{}}Multimodal Dialogue \\ Conversational Embodied Dialogue\end{tabular} & \begin{tabular}[c]{@{}c@{}}MMCoQA \cite{li2022mmcoqa} \\ ALFRED \cite{shridhar2020alfred}\end{tabular} & 200 & Accuracy \\ \cline{2-6} 
 & Space Understanding (S-5) & nuScenes & nuScenes \cite{caesar2020nuscenes} & 200 & Accuracy \\ \hline
\multirow{3}{*}{Diagnostic Evaluation} & Needle In A Haystack (N-1) & Text Needle In A Haystack & TextNeedleInAHaystack & 320 & Accuracy \\ \cline{2-6} 
 & Needle In A Haystack (N-2) & Image Needle In A Haystack & ImageNeedleInAHaystack & 320 & Accuracy \\ \cline{2-6} 
 & Image Retrieval (I-1) & Image Retrieval & GPR1200 \cite{schall2022gpr1200} & 600 & Accuracy \\ \hline
\end{tabular}%
}
\end{table*}



 \fi

\end{document}